\documentclass[12pt]{article}
\usepackage{amsmath}
\usepackage{times}
\usepackage{graphicx}
\usepackage{color}
\usepackage{multirow}

\usepackage{rotating}
\usepackage{bbm}
\usepackage{latexsym}

\usepackage{hyperref}
\usepackage{graphicx}%
\usepackage{multirow}%
\usepackage{amsmath,amssymb,amsfonts}%
\usepackage{amsthm}%
\usepackage{mathrsfs}%
\usepackage[title]{appendix}%
\usepackage{xcolor}%
\usepackage{textcomp}%
\usepackage{manyfoot}%
\usepackage{booktabs}%
\usepackage{algorithm}%
\usepackage{algorithmicx}%
\usepackage{algpseudocode}%
\usepackage{listings}%
\usepackage{caption}
\usepackage{setspace} 

\usepackage[
backend=biber,
style=apa,
citestyle=apa
]{biblatex}

\newcommand{\captionfonts}{\normalsize}

\makeatletter  
\long\def\@makecaption#1#2{%
  \vskip\abovecaptionskip
  \sbox\@tempboxa{{\captionfonts #1: #2}}%
  \ifdim \wd\@tempboxa >\hsize
    {\captionfonts #1: #2\par}
  \else
    \hbox to\hsize{\hfil\box\@tempboxa\hfil}%
  \fi
  \vskip\belowcaptionskip}
\makeatother   

\begin{document}
\hspace{13.9cm}1

\ \vspace{20mm}\\

{\LARGE\noindent Abstracted Gaussian Prototypes for `True' One-Shot Concept Learning}

\ \\
{\bf \large Chelsea Zou, Kenneth J. Kurtz}\\
{Binghamton University, State University of New York}\\
%


\thispagestyle{empty}
\markboth{}{NC instructions}
\ \vspace{-0mm}\\
%
\begin{center} {\bf Abstract} \end{center}
We introduce a cluster-based generative image segmentation framework to encode higher-level  representations of visual concepts based on one-shot learning inspired by the Omniglot Challenge. The  inferred parameters of each component of a Gaussian Mixture Model (GMM) represent a distinct topological subpart of a visual concept. Sampling new  data from these parameters generates augmented  subparts to build a more robust prototype for each concept, i.e., the Abstracted Gaussian Prototype (AGP). This framework addresses one-shot  classification tasks using a cognitively-inspired similarity metric and  addresses one-shot generative tasks through a novel AGP-VAE pipeline employing variational  autoencoders (VAEs) to generate new class variants. Results from human judges reveal that the generative pipeline produces novel examples and classes of visual concepts that are broadly indistinguishable from those made by humans. The proposed framework leads to impressive, but not state-of-the-art, classification accuracy; thus, the contribution is two-fold: 1) the system is  low in theoretical and computational complexity yet achieves the standard of 'true' one-shot learning by operating in a fully standalone manner unlike existing approaches that draw heavily on pre-training or knowledge engineering; and 2) in contrast with existing neural network approaches, the AGP approach addresses the importance of broad task capability emphasized in the Omniglot challenge (successful performance on classification and generative tasks). These two points are critical in advancing our understanding of how learning and reasoning systems can produce viable, robust, and flexible concepts based on literally no more than a single example.

\section{Introduction}\label{sec1}
\indent The ability of humans to learn novel concepts after minimal exposure to examples is an important constituent of general intelligence. Humans have the remarkable ability to quickly abstract concepts and extrapolate from one or few provided examples \parencite{lake2015human}, thereby allowing for efficient and adaptable learning and reasoning. On the contrary, most current machine learning (ML) architectures require large amounts of data to learn, massive numbers of parameters (e.g., GPT-3: 175B, AlexNet: 62.3M, VGG16: 138M), and access to externally trained models or data \parencite{hendrycks2019using, han2021pre, l2017machine, zhou2016learnware}. Hence, a key computational challenge is to understand how an intelligent system with minimal complexity and without reliance on external supports can acquire new concepts from highly limited training data \parencite{chollet2019measure}.

In this paper, we approach the challenges of one-shot learning \parencite{wang2020generalizing,kadam2020review} by developing a framework that can perform the classification and generative tasks defined by the Omniglot challenge of handwritten characters—a testbed designed to promote the study of human-like intelligence in artificial systems \parencite{lake2015human,lake2019omniglot}. In the classification task, a single image of a novel character is presented and the aim is to correctly identify another instance of that character from a choice set of characters. In the generative tasks, the goal is to create new variants of characters that are indistinguishable from human drawings. While the classification task has received much attention, there has been limited success in the attempt to achieve both types of tasks with the same model (despite an emphasis on exactly this breadth of functionality in the Omniglot challenge).

The Omniglot challenge is a highlight of the emerging field of computational cognition. Unlike traditional ML research that focuses on the narrow goal of achieving good classification performance on novel test items given a large set of labeled training data, the emphasis is instead on robust and flexible performance from  maximally sparse training data. The Omniglot challenge differs in ways that derive from perceived weakness in dominant ML approaches relative to human intelligence: 1) the ability to form a concept from a single training item rather than expansive training sets; and 2) the ability to form flexible, robust, and interpretable concepts. Accordingly, the challenge focuses on one-shot learning under a broad, multi-task interpretation of inductive concept learning that asks proposed systems not just to classify new items, but also to successfully perform a set of generative tasks. The latter includes inventing viable new instances of an individual character in an alphabet, inventing viable new characters within an alphabet, and inventing viable new characters without constraint (note: an additional proposed task involves parsing characters into ordered strokes, however we see this as less central to the thrust of the challenge). 

It is impressive that (despite early poor performance) deep learning models are now close to optimal on one-shot classification. However, the proposed solutions have serious shortcomings with regard to the letter and spirit of the challenge. In our work, we pursue an approach that hews closely to the actual formulation of the Omniglot challenge (adhering to the  challenge's emphasis on breadth of task performance) and in an important way \textit{heightens} the challenge. The originators of the Omniglot challenge took the view that new acts of learning are grounded in past 'learning how to learn' which provides constraints on the vocabulary of representation and priors to guide induction and prediction in a new domain. Consider the case of most human learning:  new learning is often informed by other known concepts that are relevant as well as by general knowledge of how the world works. This leads to a rather free interpretation of one-shot learning that permits the use of pre-training on outside data or the use of pre-established sophisticated knowledge engineering. We  instead adopt a strict interpretation of 'true' one-shot learning as learning entirely from scratch: Can a system succeed on the Omniglot challenge if it begins as a blank slate with no exposure to additional data, no prior knowledge, and no pre-established symbol system?

An important intellectual thread embedded in the Omniglot challenge is a question at the heart of computational cognition: Are structured (also referred to as compositional, symbolic, or causal) representations  necessary to learn and reason with concepts in a sophisticated manner? We attempt to avoid a  black-and-white approach to this question. Rather than committing to massive data and architectures to  estimate models that yield high performance in tightly constrained evaluations or assuming the availability of powerful symbolic systems prior to new learning, we investigate design principles to construct quasi-structured representations of a concept that start with nothing more than a single training item.

\begin{figure*}[t]
  \centering
\includegraphics[width=\textwidth]{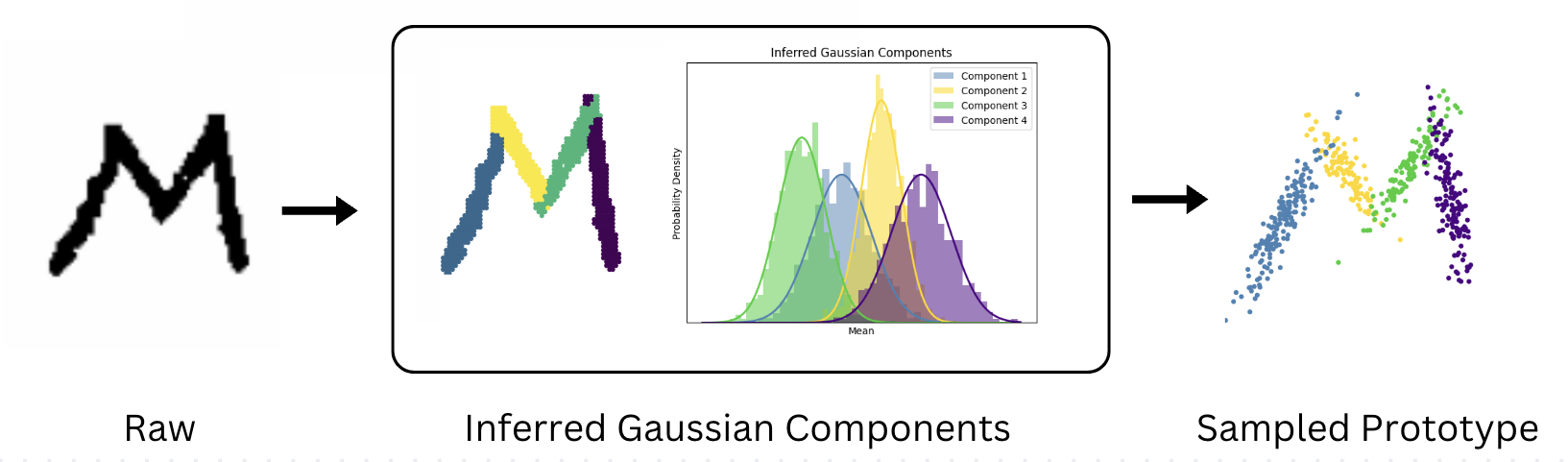} 
  \caption{The raw image is shown on the left. The inferred clusters of the GMM are shown in the middle. The newly generated abstracted Gaussian prototype (AGP) is sampled from the inferred parameters.}
  \label{fig:prototype}
\end{figure*}

To address both classification and generative tasks of the Omniglot challenge, we propose the \textit{Abstracted Gaussian Prototype} (AGP) which leverages Gaussian Mixture Models (GMMs) to flexibly model visual concepts of handwritten characters. This is accomplished by capturing character subparts as inferred Gaussian components that flexibly represent 'what should be where' for a concept based on a single example. GMMs are unsupervised clustering algorithms that represent data as a finite sum of Gaussian distributions \parencite{duda1973pattern, yu2015gaussian, liang2022gmmseg, reynolds2009gaussian}. A key insight driving our approach is viewing the pixels of an image as instances in the domain of an individual character. We use GMM-based clustering to model the topological subparts of each class in terms of its unique distribution. Based on the parameters underlying each component's distribution, the GMMs are then used to generate additional model-consistent pixels that augment the subpart representations. Finally, the collective ensemble of these generated subparts form what we refer to as the AGP, see Figure \ref{fig:prototype}. 

AGPs provide a way to extrapolate beyond the constraints of a single available instance based on: 1) the clustering functionality that imparts a quasi-structural analysis of the raw input into underlying parts with relative locations; and 2) the generative functionality that enhances the item encoding by extrapolating its underlying model. This is a promising intermediate strategy between the problem of completely lacking a structural representation and that of being overly structured in the sense of requiring the overhead of a built-in representational vocabulary and being tied to fixed expectations. In general, AGPs offer a method for generative image segmentation. The resulting higher-level representation (AGP) is successfully induced from a single example and supports a range of conceptual tasks. AGPs are characterized by two elements: 1) probabilistic capture of data substructures, and 2) parameterized generative modeling. The first element allows the system to accurately model diverse concepts and robustly handle noise. The second element allows the creation of sufficiently flexible representations of each concept.

To achieve one-shot classification, we leverage the generalized class representation of AGPs in combination with a cognitively-inspired metric to assess the similarity between AGPs. A classic and highly influential psychological theory of human similarity judgement is a set-theoretic approach known as the contrast model \parencite{tversky1977features}. This similarity metric employs an asymmetric weighting of the number of common and distinctive features between being compared. In the context of the one-shot classification task, the features to be evaluated are pixel intersections between AGP representations of the characters (further details below), so a classification decision from a choice set is made based on the highest similarity score to the target item. For the challenging generative tasks in the Omniglot challenge requiring the invention of new character types or tokens, we build on the approach outlined above to devise an AGP-VAE pipeline to create new classes. First, we synthetically generate a diverse training set of AGPs from each provided character. Then, we employ a variational autoencoder (VAE) \parencite{kingma2013auto}, a type of generative neural network architecture, to learn a continuous latent space that encapsulates a probabilistic distribution over different classes of generated AGP train sets. This novel AGP-VAE pipeline has the ability to interpolate between subparts of the discrete prototypes of the AGPs by sampling from a global feature space that encapsulates the local features of different classes. Instead of relying on the independent models of each class, we are able to represent the available classes in a subspace that can be sampled to produce novel character classes that conform to the distributional characteristics of provided data. In sum, the key design principles of our contribution are as follows:
\begin{enumerate}
  \item We formulate the framework of an AGP as a generative image segmentation method that provides a higher-level representation from a single character instance by applying a clustering approach at the pixel level.
  \item We make novel use of a cognitively inspired similarity metric between AGP representations to perform one-shot classification.
  \item We employ a novel AGP-VAE pipeline for generative tasks to create new visual concepts. 

\end{enumerate}

Our framework provides a flexible and robust way to handle both classification and generative tasks. The system learns and performs these tasks without reliance on pre-training or a pre-formed symbolic system. Further, the approach is  highly transparent and relies on a small number of clear design principles that are novel applications of established computational constructs. 

\section{Prototypes in Human Concept Learning}
Concepts are mental representations acquired through inductive learning processes that are used to categorize and reason about objects, events, and relational situations \parencite{gregory2002big}. Psychologists have emphasized that concepts are formed and represented through the integration and combination of simpler parts known as features or attributes \parencite{schyns1998development,farhadi2009describing}. In the concepts and categories literature in cognitive psychology, a prominent account is prototype theory in which the mental representations of categories consist of a stored summary of the central tendency of feature values across observed members of a category \parencite{rosch1973natural,hampton2006concepts,posner1968genesis, minda2011prototype}. The prototype of a concept is a statistical average or an abstraction of all instances observed as members of a category. Categorization then operates as a matter of finding the best match among candidate prototypes which supports graded membership and classification decisions based on family resemblance rather than logical rules \parencite{rosch1978principles, mervis1981categorization}. This approach has contrasted sharply in the psychological literature with exemplar-based similarity approaches that eschew explicit abstraction in favor of storing labeled instances \parencite{nosofsky1988exemplar,nosofsky1986attention}. The main advantage of prototype theory is its broad computational and psychological plausibility: prototypes are economical in terms of representational and processing requirements, and they capture the intuition that the generic meaning underlying a category is represented explicitly and independently of its members.

In our framework, we adapt a form of prototype theory to address the Omniglot challenge. Prototypes are not necessarily formed by storing the centroid of category members but can instead be realized implicitly through statistical learning or neural network architectures like simple linear associators or auto-associative systems which adapt their weights to capture the statistical regularities of a domain \parencite{biehl2016prototype,rosenblatt1958perceptron,ruck1990feature}. 
To be clear, prototypes are nearly always a matter of abstracting a summary representation or model from individual cases. By contrast, our approach to one-shot learning transforms a single example into a prototype by taking a particular configuration of pixels as the basis for a set of probabilistic clusters that imply an underlying distribution with a generative capacity. The crux of our approach lies in the formation of an ensemble of augmented subparts achieved by using the parameters inferred from the Gaussian components of the GMMs. The abstracted Gaussian prototype is an abstraction from a single case in order to create its own subparts via clustering, establish the distributional central tendency and variability of its subparts,  inherently capture spatial relational properties between the subparts, and further populate the prototype with distribution-consistent generated pixels. The result of this process can be seen as similar to a summary representation of the central tendency across varying training examples with the additional quasi-structural aspect of an underlying model that captures the form and location of subparts (without any explicit propositional or symbolic representation).

Just as the prototype theory of human categorization relies on similarity to the stored prototype in order to determine likelihood of category membership, we invoke a psychological similarity metric for classification relative to AGPs. Tversky's (1977) model of similarity proposed that individuals assess similarity by considering the number of featural differences and commonalities between items along with an additional design principle of highlighting (via greater weighting) the importance of the differences. The Tversky index is given by:
\begin{equation}
   T(A, B) = \frac{|A \cap B|}{|A \cap B| + \alpha |A \backslash B| + \beta |B \backslash A|} 
\end{equation}

\noindent where $|A \cap B|$ is the size of the intersection of sets $A$ and $B$,
$|A \backslash B|$ is the size of the set difference of $A$ without $B$, 
$|B \backslash A|$ is the size of the set difference of $B$ without $A$, and
$\alpha, \beta$ are non-negative parameters controlling the weight given to differences in the two sets. We employ a slightly simpler metric that has only one parameter to control the magnitude of the penalty to the set differences in comparing the intersections of AGPs, described in Section \ref{simmetric}.

We do not claim the approach to be a truly blank slate learner. Learning depends on inductive bias and in our case, the biases are explicit and intentionally lightweight: characters are treated as 2D point sets of foreground pixels and their structure is summarized by a small number of Gaussian components (capturing local stroke-like regularities). Similarity is computed with a small tolerance to account for minor shifts. These assumptions inject domain knowledge but avoid learned priors from massive amounts of external data (which is typically used by more programmatic stroke-based models). We view this as a simpler and more interpretable middle ground between heavily structured symbolic models and heavily pretrained neural models. One might ask why this constraint is worth imposing, given that humans learn new concepts by building on prior knowledge and that foundation models achieve strong results through pre-training. We see several reasons. 1) If pre-training is assumed necessary for good performance, it is informative to establish what can be achieved without it. 2) A simpler system with fewer dependencies is preferable from both scientific and engineering standpoints—success and failure can be clearly attributed to stated design principles rather than to opaque upstream learning. 3) The premise of one-shot learning is performance under minimal information, and leveraging extensive prior training sidesteps rather than addresses that constraint. Our aim is not to reject foundation-model approaches but to complement them by exploring what transparent, low-complexity structure alone can accomplish.

\section{Related Works}

\subsubsection{One-Shot Classification:} Many neural-based models have been successful at one or few-shot classification tasks \parencite{finn2017model, santoro2016meta, salakhutdinov2012learning}. For instance, Siamese Neural Networks involve twin sub-networks sharing the same parameters that are trained to learn embeddings capturing the similarity or dissimilarity between pairs of instances \parencite{koch2015siamese, chicco2021siamese}. Another approach through Prototypical Networks learn a representative prototype for each class based on the mean of the embeddings in the latent space and classify according to these distances \parencite{snell2017prototypical}. Similarly, Matching Networks work by employing an attention mechanism on embeddings of the labeled set of instances to forecast classes for unlabeled data \parencite{vinyals2016matching}. However, all of these neural-based classification approaches require an initial training phase for the network to learn a general understanding of the task. Furthermore, they are incapable of addressing generative tasks. Our proposed approach, on the other hand, offers a direct way for both classification and generative tasks to learn specific concepts from one, and only one, shot without background training on other data. 

\subsubsection{One-Shot Generation:} One recent approach introduces one-shot generative domain adaptation (GenDA) using pre-trained Generative Adversarial Networks (GANS) \parencite{yang2023one, goodfellow2014generative}. GenDA designs an attribute classifier that guides the generator to optimally capture representative attributes from a single target image, in turn, synthesizing high-quality variant images. However, this approach relies on source models that are pre-trained on large-scale datasets such as FFHQ \parencite{ffhq} and Artistic-Faces dataset \parencite{artistic}. 

\subsubsection{Bayesian Models: } Significant progress has been made to address both one-shot classification and generative tasks through Bayesian implementations, such as the Object Category Model  \parencite{fei2006one} involving parametric representations of objects and prior knowledge when faced with minimal training examples. These Bayesian principles are manifested in the approach proposed by the original authors for the Omniglot dataset \parencite{lake2011one}. In this system, a stroke model learns part-based representations from previous characters to help infer the sequence of latent strokes in new characters. An extension of this work introduces Bayesian Program Learning (BPL), which learns a dictionary of sub-strokes and probabilistically generates new characters by constructing them compositionally from constituent parts and their spatial relationships \parencite{lake2015human}. BPL and the stroke model, however, requires the model to learn from stroke-data trajectories in order to extract and store a dictionary of primitive parses at the sub-stroke level. While this model first requires information from live-drawings, the approach may be unfeasible when temporally labeled sequential stroke data is not accessible. In contrast, our approach uses generative image segmentation to directly infer the sub-strokes of the characters using GMMs which allows our model to learn purely from a single raw image.

\subsubsection{Gaussian Splatting}
Our representation is related to Gaussian splatting, where an image is expressed as a superposition of Gaussian primitives that are rasterized onto a grid through a rendering step. In contrast, our use of Gaussian is explicitly probabilistic: we fit a GMM to the set of foreground pixel coordinates from a single binary instance and use the fitted density to 1) sample additional coordinates for prototype augmentation and 2) compare instances using a set-based matching score. Rather than defining the image as a continuous intensity field produced by summing Gaussian contributions, we define a distribution over 2D coordinates whose samples yield a point-set. This probabilistic view is convenient in the one-shot setting because it provides a simple, identifiable fitting procedure from a single image and a principled mechanism for generating controlled within-class variability.

\subsubsection{Limitations of related works}
In essence, existing  approaches rely on either pre-training and/or built-in representational systems that fall outside of the scope of our stricter interpretation of the challenge; and many   do not address the breadth of tasks in Omniglot. For instance, \parencite{b1} involves pre-training as part of its methodology to achieve one-shot generalization, which is a common approach in deep generative models to understand and generalize from very few examples. In \parencite{b2}, the concept of the neural statistician implies learning statistical representations from data, which also involve pre-training to capture the statistical properties across different contexts or datasets. In \parencite{b6}, given its neuro-symbolic approach, this paper uses built-in knowledge/representational languages which involves more explicit and expensive forms of knowledge representation. Finally, in \parencite{b3,b4,b5}, these papers use highly complex methodologies and low transparency architectures like GANS and hierarchical approaches that attempt to model data at multiple levels of abstraction. While the performance of these works offer solutions to few shot learning, our work attempts to capture a different focus. Our system is trained on nothing except the presented single characters for each task.

\section{Background}
In this section, we provide the mathematical background underlying GMMs and VAEs.

\subsection{Gaussian Mixture Models}

A Gaussian mixture model (GMM) represents a density over $d$-dimensional observations $x\in\mathbb{R}^d$
as a combination of $k$ Gaussian components \parencite{duda1973pattern}:
\begin{equation}
p(x \mid \theta) \;=\; \sum_{i=1}^{k} \pi_i \, \mathcal{N}(x \mid \mu_i, \Sigma_i),
\label{eq:gmm}
\end{equation}
where $\pi_i \ge 0$ and $\sum_{i=1}^{k}\pi_i = 1$, and $\theta=\{(\pi_i,\mu_i,\Sigma_i)\}_{i=1}^{k}$. The multivariate Gaussian density is
\begin{equation}
\mathcal{N}(x \mid \mu,\Sigma) \;=\; \frac{1}{(2\pi)^{d/2}|\Sigma|^{1/2}}
\exp\!\left(-\frac{1}{2}(x-\mu)^{\top}\Sigma^{-1}(x-\mu)\right),
\label{eq:mv_gaussian}
\end{equation}
where $\mu\in\mathbb{R}^{d}$ and $\Sigma\in\mathbb{R}^{d\times d}$. In practice, $\theta$ is commonly estimated by maximum likelihood using the expectation maximization (EM) algorithm.

\subsection{Variational Autoencoders}

Variational autoencoders (VAEs) are neural networks characterized by an encoder-decoder architecture. Unlike traditional autoencoders \parencite{bank2023autoencoders}, VAEs are equipped with a probabilistic framework based on variational inference. This framework employs networks to approximate otherwise computationally intractable posterior distributions, enabling the model to learn continuous representations of discrete input classes \parencite{kingma2013auto, kingma2019introduction}. This latent space facilitates the generation of diverse and novel samples, thereby making VAEs a versatile tool for tasks such as image generation.

\subsubsection{Encoder:}
The encoder of a VAE maps input data $x \in \mathbb{R}^d$ to a latent variable $z \in \mathbb{R}^J$, where $J$ is the dimensionality of the latent space. The encoder defines an approximate posterior distribution with a diagonal covariance structure:
\begin{equation}
   q_{\phi}(z \mid x) = \mathcal{N}\!\left(z;\, \mu_{z}(x),\, \mathrm{diag}(\sigma_{z}(x)^2)\right),
\end{equation}
where $\mu_{z}(x) \in \mathbb{R}^J$ and $\sigma_{z}(x) \in \mathbb{R}^J$ are the mean vector and standard deviation vector of the approximate posterior, learned by the encoder with parameters $\phi$.

\subsubsection{Sampling with Reparameterization Trick:}
To obtain a sample $z \sim q_{\phi}(z \mid x)$ while maintaining differentiability for backpropagation, the reparameterization trick is used:
\begin{equation}
    z = \mu_z(x) + \sigma_z(x) \odot \epsilon, \quad \epsilon \sim \mathcal{N}(0, I_J),
\end{equation}
where $I_J$ is the $J$-dimensional identity matrix and $\odot$ denotes element-wise multiplication.

\subsubsection{Decoder:}
The decoder with parameters $\theta$ maps a latent sample $z$ back to the original feature space. The conditional likelihood is defined as:
\begin{equation}
    p_{\theta}(x \mid z) = \mathcal{N}\!\left(x;\, \mu_x(z),\, \mathrm{diag}(\sigma_x(z)^2)\right),
\end{equation}
where $\mu_{x}(z) \in \mathbb{R}^d$ and $\sigma_{x}(z) \in \mathbb{R}^d$ are the mean and standard deviation of the reconstructed data. The reconstruction is given by $\hat{x} = \mu_x(z)$.

\subsubsection{Loss Function:}
The training objective for VAEs is to maximize the Evidence Lower Bound (ELBO):
\begin{equation}
   \mathcal{L}(\phi, \theta) = 
    \mathbb{E}_{q_{\phi}(z \mid x)}\!\left[\log p_{\theta}(x \mid z)\right] - D_{\mathrm{KL}}\!\left(q_{\phi}(z \mid x) \,\|\, p(z)\right),
\end{equation}
where the first term is the expected log-likelihood of $x$ given $z$ (the reconstruction term) and the second term is the Kullback--Leibler (KL) divergence between the approximate posterior and the prior. The KL divergence acts as a regularization term that encourages the approximate posterior to remain close to the prior $p(z)$. In practice, the loss minimized during training is $-\mathcal{L}(\phi, \theta)$. In a standard VAE, the prior is assumed to be a multivariate standard normal distribution, $p(z) = \mathcal{N}(z;\, 0, I_J)$. Under this assumption and the diagonal Gaussian encoder, the KL divergence admits a closed-form expression:
\begin{equation}
    D_{\mathrm{KL}}\!\left(q_{\phi}(z \mid x) \,\|\, p(z)\right) = 
    \frac{1}{2} \sum_{j=1}^{J} \left( \mu_j^2 + \sigma_j^2 - \log \sigma_j^2 - 1 \right),
\end{equation}
where $\mu_j$ and $\sigma_j$ are the $j$-th components of $\mu_z(x)$ and $\sigma_z(x)$, respectively.

\section{Methods}
In this section, we describe and formalize our approach to the classification and generative tasks in the Omniglot Challenge. The Omniglot dataset consists of 1623 hand-written characters taken from 50 different alphabets, with 20 examples for each class \parencite{lake2019omniglot}. All code, question sets, and data can be accessed at \url{https://github.com/bosonphoton/AbstractedGaussianPrototypes}

\subsection{Classification Task}
In a one-shot classification task, there is a set of $N$ classes 
$\mathcal{C} = \{c_1, c_2, \ldots, c_N\}$, with exactly one available 
instance per class. The set of these single instances is denoted 
$\mathcal{X} = \{x_1, x_2, \ldots, x_N\}$, where $x_i$ is the instance 
corresponding to class $c_i$. Given an unseen query instance $q \notin 
\mathcal{X}$ belonging to some class in $\mathcal{C}$, the goal of one-shot 
classification is to correctly determine its class label $c_q \in \mathcal{C}$.

Our classification function proceeds in two steps: (1) Abstracted Gaussian 
Prototype (AGP) generation, and (2) a cognitively inspired similarity metric 
to determine classification. First, a GMM is used to separately model each 
instance, where each mixture component is assumed to represent a distinct 
spatial subpart of the character. We then generate enriched populations by 
sampling from the fitted components; the resulting ensemble of generated 
subparts forms the AGP, a higher-level representation for each class, denoted 
$\mathcal{P}_i$ for instance $x_i$. The query $q$ undergoes the same AGP 
generation process, yielding $\mathcal{P}_q$. Using these prototypes, we 
compute the similarity between $\mathcal{P}_q$ and each $\mathcal{P}_i$ for 
$i \in \{1, \ldots, N\}$ using a metric based on Tversky's contrast model 
\parencite{tversky1977features}. The class whose prototype produces the highest 
similarity score is selected as the predicted class of $q$. The details of 
each step are formalized below.

\subsubsection{Abstracted Gaussian Prototype (AGP) Generation} 
\label{gmmstep}

Each character instance is a binary image on an $H \times W$ grid. We convert 
an image $I$ into a set of foreground pixel coordinates
\[
S(I) \;=\; \{s_j \in \mathbb{R}^2\}_{j=1}^{m}, \qquad s_j = (u_j, v_j),
\]
where $s_j$ indexes the locations of ``on'' pixels with value $1$ (ink), and 
$m$ is the number of foreground pixels in the instance. We model these 
foreground coordinates as i.i.d.\ samples from a $k$-component GMM
\[
p(s \mid \theta) \;=\; \sum_{i=1}^{k} \pi_i \, 
\mathcal{N}(s \mid \mu_i, \Sigma_i),
\]
with parameters $\theta = \{(\pi_i, \mu_i, \Sigma_i)\}_{i=1}^{k}$, mixture 
weights $\pi_i \ge 0$, and $\sum_{i=1}^{k} \pi_i = 1$.

Given an instance $I$, we estimate $\theta$ from $S(I)$ via the EM algorithm. 
Each mixture component is interpreted as a spatial ``part'' that captures a 
locally coherent region of the character (e.g., a stroke fragment), and $k$ is 
a hyperparameter controlling the granularity of this decomposition.

After fitting the GMM, we generate a component-wise prototype by sampling 
additional coordinates from each Gaussian component. Specifically, for each 
component $i \in \{1, \dots, k\}$ we draw
\[
p_i \;=\; \{s^{(i)}_t\}_{t=1}^{n_i}, \qquad 
s^{(i)}_t \sim \mathcal{N}(\mu_i, \Sigma_i),
\]
where $n_i = \lfloor n \cdot \pi_i \rceil$ is the number of samples allocated 
to component $i$, proportional to its mixture weight, and $n$ is the total 
number of generated samples per instance.

The resulting set
\[
\mathcal{P}(I) \;=\; \{p_1, p_2, \ldots, p_k\}
\]
constitutes the component-wise prototype for the instance, and its union 
$\bigcup_{i=1}^{k} p_i$ is rasterized back onto the pixel grid when an image 
representation is needed. During rasterization the model samples with replacement, so cases in which multiple sampled points map to the same pixel will just simply collapse to a single ``on'' pixel.

\subsubsection{Similarity Metric} \label{simmetric}

The similarity metric is computed between the query prototype $\mathcal{P}_{q}$ and each available prototype $\mathcal{P}_{i}$. We base our similarity metric on the Tversky index with the following equation:
\begin{equation}
      S(A,B) = \ |A \cap B| - \beta \ |A \triangle B|
      \label{equation12}
\end{equation}
\noindent where $A \triangle B = (A\backslash B) \cup (B\backslash A)$ is the symmetric difference representing the non-intersections and $\beta > 1$ is a weight hyperparameter that ensures a larger penalty of this difference.
Let $A$ be the rasterized set of foreground pixel coordinates for $\mathcal{P}_{q}$, and $B$ the corresponding set for $\mathcal{P}_{i}$.
We introduce a small tolerance radius $r > 0$ to allow matches under minor pixel jitter.
Define the indicator that a pixel $a\in A$ has a match in $B$ within radius $r$:
\begin{equation}
\operatorname{isMatched}(a,B;r) \;=\;
\begin{cases}
1, & \text{if } \exists\, b \in B \text{ such that } \lVert a - b \rVert_2 \le r,\\
0, & \text{otherwise,}
\end{cases}
\label{eq:isMatched}
\end{equation}
where $a = (x_a, y_a)$ and $b = (x_b, y_b)$ are pixel coordinates.
The number of matched pixels from $A$ to $B$ is
\begin{equation}
\operatorname{Matched}(A,B;r) \;=\; \sum_{a\in A} \operatorname{isMatched}(a,B;r),
\label{eq:matched}
\end{equation}
and the number of unmatched pixels from $A$ to $B$ is
\begin{equation}
\operatorname{Unmatched}(A,B;r) \;=\; \sum_{a\in A}\Big(1-\operatorname{isMatched}(a,B;r)\Big).
\label{eq:unmatched}
\end{equation}
We compute matched and unmatched pixels in both directions. The overlap is taken as the minimum of the two matched counts, which prevents inflated scores when one set is much larger than the other. The penalty sums the unmatched counts in both directions:
\begin{equation}
\operatorname{Overlap}(A,B;r) \;=\; \min\!\Big(\operatorname{Matched}(A,B;r),\;
\operatorname{Matched}(B,A;r)\Big),
\label{eq:overlap}
\end{equation}
\begin{equation}
\operatorname{TotalUnmatched}(A,B;r) \;=\; \operatorname{Unmatched}(A,B;r) + \operatorname{Unmatched}(B,A;r).
\label{eq:total_unmatched}
\end{equation}
Our similarity score is then
\begin{equation}
S(A,B) \;=\; \operatorname{Overlap}(A,B;r) \;-\; \beta\, \operatorname{TotalUnmatched}(A,B;r),
\label{eq:similarity}
\end{equation}
where $\beta > 1$ controls the penalty on non-overlap.
\noindent Finally, the $\mathcal{P}_{i}$ with the highest similarity score with $\mathcal{P}_{q}$ is deemed to be of the same class:
\begin{equation}
c_{q} \;=\; \underset{i \in \{1,\dots,N\}}{\arg\max}\; S(A_q, B_i),
\label{eq:classification}
\end{equation}
where $B_i$ denotes the foreground pixel set of $\mathcal{P}_{i}$.
In order to improve the quality of the similarity score, we perform min-max scaling on each image pair and shift them to a center grid to align them as best as possible. For the similarity computations in the classification tasks, the query is also evaluated under eight spatial transformations: translations in the four cardinal directions, and rotations of $\pm 15^{\circ}$ and $\pm 25^{\circ}$, and the transformation yielding the highest similarity is selected.

\noindent \textbf{Parameter Estimation:} The number of clusters \( k \) in the GMM and the weight parameter \( \beta \) in equation \eqref{equation12} were optimized using a grid-search approach to maximize classification accuracy on a held-out validation set. For \( k \), we varied the number of components from 6 to 10 in increments of 1, and for \( \beta \), we tested values in the range \([1.0, 2.0]\) with a step size of 0.2. The chosen values (\( k = 10 \), \( \beta = 1.4 \)) corresponded to the highest average classification accuracy across 100 iterations. The theoretical interpretation of these parameters relates directly to the structure and granularity of the concepts being modeled. A larger \( k \) allows the model to represent finer-grained subparts of each character for capturing detailed structural variability. Conversely, a smaller \( k \) enforces a coarser representation. The parameter \( \beta \) balances the contributions of overlapping and non-overlapping pixels in the similarity computation. A larger \( \beta \) emphasizes differences and penalizes mismatched regions more heavily, while a smaller \( \beta \) places greater weight on commonalities.



\begin{figure*}[t]
  \centering
\includegraphics[width=\textwidth]{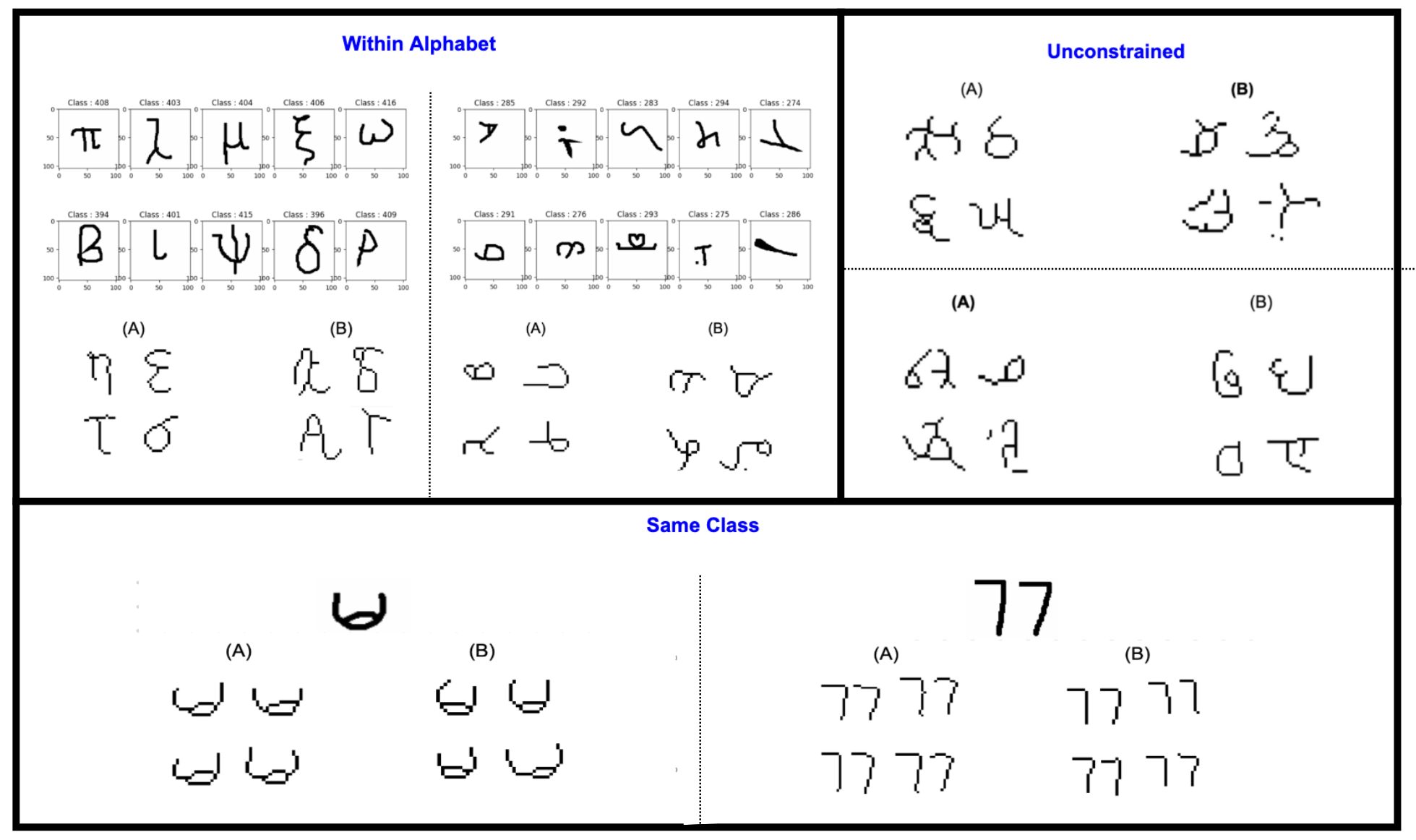} 
  \caption{Visual Turing tests of the output characters generated from our AGP-VAE pipeline. The set of characters drawn by our model is B for within alphabet, A for same class, and B for unconstrained.}
  \label{fig:generated}
\end{figure*} 

\subsection{Generative Tasks}
The generative tasks in the Omniglot challenge are as follows: 
\begin{enumerate}
    \item Generating new exemplars of a particular class (character).
    \item Generating new classes consistent with a particular alphabet given a starting set.
    \item Generating entirely new classes (unconstrained).
\end{enumerate}

For Task (1), only a single instance is used for the entire task. For Task (2), one instance per class from a given alphabet is allowed. Following \parencite{lake2019omniglot}, we use ten different instances. Similarly, Task (3) uses the same technique as Task (2) except that classes are randomly sampled across alphabets (note: this seems to be the best interpretation of "unconstrained" given that the model must be given some type of input). Overall, the approach between these three tasks is similar and the only difference is in the starting instance(s) that are used. 

There are three steps to generating new variations of exemplars from single instances. First, we synthetically increase the amount of training data by generating a larger number of AGPs per class. A key feature of this approach is that a range of prototypes with varied manifestations of abstraction can be generated by specifying a range of different numbers of components in the GMM for each class. This increases variation at the subparts level to diversify the training set. Second, we train a VAE across all the synthetic data to learn a continuous space of prototypes derived from the starting set. Lastly, we use a post-processing topological skeleton technique \parencite{lee1994building,zhang1997fast} to refine the generated outputs. This layer denoises the  reconstructed outputs of the VAE to ensure quality stroke images of the new class variants. The following sections describe each step in detail, and the pseudocode is shown below in Algorithm \ref{alg:generative}.

\subsubsection{AGP Training Set}

We synthetically increase the training data by generating multiple AGP samples per class.
For each class $i\in\{1,\dots,N\}$ and each mixture size $k \in K$ drawn from a set of candidate component counts $K$, we fit a GMM to the foreground coordinates of the available instance,
sample a prototype coordinate set, and rasterize it to a binary image. We denote the resulting rasterized AGP image by $\tilde{x}_{i,k,s} \in \{0,1\}^{H\times W}$,
where $s$ indexes repeated samples.

The full augmented dataset used to train the VAE is the union of all rasterized AGP images:
\begin{equation}
\Phi \;=\; \bigcup_{i=1}^{N} \bigcup_{k \in K} \bigcup_{s=1}^{D_k} \{\tilde{x}_{i,k,s}\}.
\end{equation}
In our pipeline we use $K=\{6,7,8,9,10\}$ and generate $D_k$ samples per $k$ such that the total number of AGPs per class is $D=\sum_{k\in K}D_k=500$
(i.e., $D_k=100$ for each $k$).


\begin{singlespace} 
\begin{algorithm}
\caption{Generating New Variants}
\label{alg:generative}
\textbf{Input:} \\
$\mathcal{X} \gets$ set of single instances from $N$ classes\\
$K \gets$ range of Gaussian components \\
GMM, VAE $\gets$ trainable models \\
Skel $\gets$ topological skeletonization function \\
\textbf{Output:} \\
$v \gets$ final generated variant \\
\begin{algorithmic}[1]
\State $\Phi \gets \emptyset$
\For{$i$ in $N$}
    \For{$k$ in $K$}
        \State fit $GMM_{i,k}$ to foreground coordinates of $x_i$ using $k$ components
        \For{$s = 1$ \textbf{ to } $D_k$}
            \State $\mathcal{P}_{i,k,s} \gets$ sample prototype points from $GMM_{i,k}$
            \State $\tilde{x}_{i,k,s} \gets$ rasterize $\mathcal{P}_{i,k,s}$ to a binary image
            \State $\Phi \gets \Phi \cup \{\tilde{x}_{i,k,s}\}$
        \EndFor
    \EndFor
\EndFor
\State train VAE($\Phi$)
\State $z \gets$ sample from VAE latent space
\State $\hat{x} \gets$ decode $z$
\State $v \gets$ Skel($\hat{x}$)
\end{algorithmic}
\end{algorithm}
\end{singlespace}
\subsubsection{VAE Interpolation}
After generating the prototype training sets $\Phi$ the next step is to create continuous variations among these prototypes. To accomplish this, a VAE is trained across this enumeration of all prototype train sets to learn a latent space representation that captures the underlying structures of these abstracted prototypes. The latent variables $z$ are sampled manually to encourage semantic mixing between prototypes which are then decoded into the reconstructed variant images. For our generated outputs, we specify a convolutional VAE with the following details. The encoder of our model consists of two convolutional layers: a 32-filter 3x3 convolution followed by a 64-filter 3x3 convolution, both with a stride of 2 and ReLU activation. The decoder reshapes the latent vector to a 7x7x32 tensor, upsampled using two transposed convolutional layers with 64 and 32 filters respectively, both with 3x3 kernels, a stride of 2, and ReLU activation. The final layer is a transposed convolution with a single filter, a 3x3 kernel, and stride of 1, outputting the reconstructed image. For training, the model uses the Adam optimizer \parencite{kingma2014adam} with a learning rate of 1e-4. The loss function is computed as a combination of binary cross-entropy and the KL divergence between the learned latent distribution and the prior distribution. The model is trained using a batch size of 32 for 50 epochs.

\subsubsection{Topological Skeleton Refinement}
The final step in this pipeline is a post-processing technique based on the work of \parencite{lee1994building,zhang1997fast} on topological skeletons. Skeletonization is used often in image processing and computer vision to reduce the thickness of binary objects to one-pixel-wide representations while preserving the topological properties of objects. This step refines the reconstructed output images generated by the VAE and emphasizes the stroke-like properties of Omniglot characters. After each reconstructed image from the VAE is skeletonized, the final result is a collection of generated variants of characters for each task.

\section{Results}
\subsection{Classification Tasks}

The classification accuracy of the proposed approach is evaluated based on the number of correct responses averaged over 1000 trials in both 5-way and 20-way one-shot tasks. We test this in the context of unconstrained classes independent of alphabets, as well as a more challenging within-alphabet classification task. We compared our Tversky-inspired similarity metric against a simple nearest-neighbor baseline computed as the mean pointwise Euclidean distance between corresponding stroke coordinates. Because each stimulus is represented as a normalized, equal-length sequence, this baseline provides a direct, interpretable measure of average geometric displacement in the underlying coordinate space. In contrast, other similarity metrics such as cosine similarity is a poor fit here, since with normalized inputs it primarily reflects angular alignment in a high-dimensional embedding rather than capturing pointwise spatial misalignment. And other image similarity metrics like SSIM targets natural-image appearance statistics like luminance, contrast, texture, etc. that are absent in our line-based strokes, making it an unnecessary baseline for this setting. Results are summarized in Table \ref{tab:baseline}. For broader context, we also report the performance of BPL, which achieves 97.7\% (3.3\% error rate) on 20 way one-shot classification. While BPL achieves higher accuracy, it operates under different modeling assumptions and computational costs; we include it as a reference point for the ceiling performance achievable with full generative program learning.

\begin{table}[t]
    \small
    \centering
    \caption{Average Classification Scores}
    \begin{tabular}{cccc}
         &  Random Chance &  Euclidean Distance & \textbf{Ours} \\
         5-Way Unconstrained &  20\% &  34.5\% & \textbf{95.1\%} \\
         5-Way Within &  20\% &  31.2\% & \textbf{86.6\%} \\       
         20-Way Unconstrained &  5\% &  19.3\% & \textbf{84.2\%} \\
         20-Way Within &  5\% &  8.9\% & \textbf{71.0\%} \\
         
    \end{tabular}
    \label{tab:baseline}
\end{table}



\subsection{Generative Tasks}
\begin{figure}[t]
\centering
\includegraphics[width=1\columnwidth]{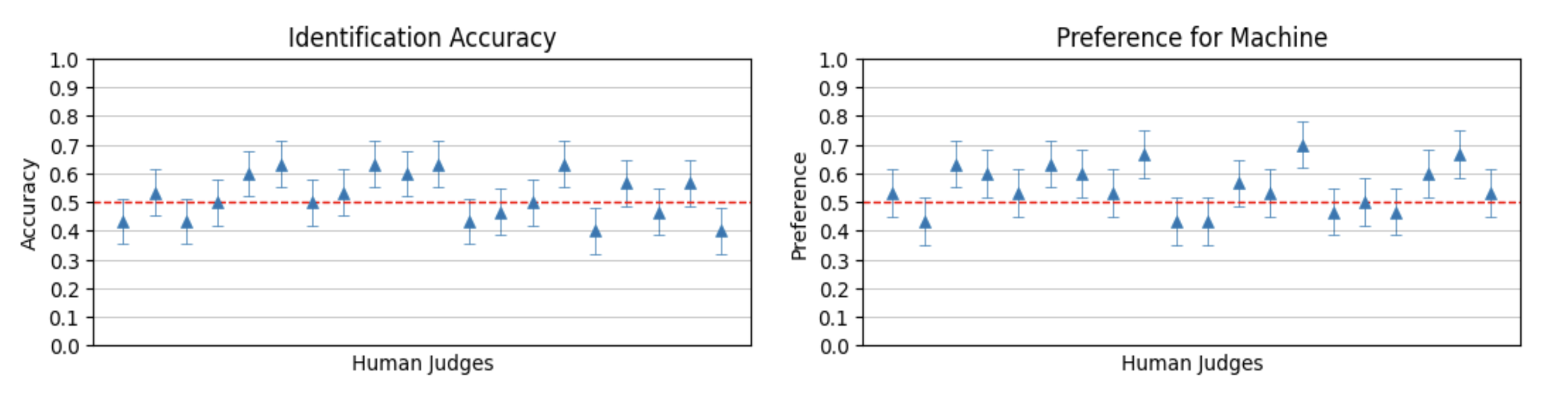}
  \caption{Each marker reveals the evaluation scores of a human judge averaged across all 30 sets of comparisons. The ideal performance is when judges cannot distinguish between human and machine outputs, which translates to random chance of 50 percent, as indicated by the dashed red line. The error bars represent the standard deviation from the mean.}
  \label{fig:individual}
\end{figure}

A "visual Turing test", as described in \parencite{lake2013one} is used to assess the quality of the generative outputs of the model. In this test, a set of characters produced by a human is displayed next to a set produced by the model. Human judges then try to identify which set was drawn by a human and which set was generated by the model, see Figure \ref{fig:generated} and Appendix \ref{app}. Our generative approach is evaluated based on the identification accuracy of  20 human judges recruited online. Since each judge is shown two images at each trial (one generated by the model and one drawn by a human), random chance is 50 percent, which is the ideal performance since it indicates that the judges find it difficult to distinguish between characters produced by the human and the model. In contrast, the worst-case performance is 100 percent, where the judges can perfectly identify that the character were drawn by either the machine or the human. Ten question sets with four instances from the human and four instances from our model were created for each of the three tasks (total of 30 sets). Additionally, we asked follow-up questions after displaying each set of images to probe whether the machine's outputs could potentially surpass the quality of human generated characters. These questions were phrased as  follows: (1) ``Which set represents a better job of making four new characters that fit the given alphabet?'', (2) ``Which set represents a better job of making four new examples of the given character?'', and (3) ``Which set represents a better job of creating four new characters?''. To assess whether the identification accuracy and preference results are reliably different from chance, we conducted statistical tests using one-sample t-tests. For identification and preference accuracy, we tested the null hypothesis where identification and preference accuracy is random, against the alternative hypothesis that the identification and preference accuracy is not random. Averaged across all tasks, the identification accuracy was (M = 52.33\%, SD = 8.17\%, Min = 40.00\%, Max = 63.33\%), p = 0.22. For the preference scores, there was a statistically significant difference (M = 55.33\%, SD = 8.27\%, Min = 43.33\%, Max = 70.00\%), p = 0.01. The results are summarized in Table \ref{tab:descriptives}. The key finding is that the identification accuracy scores show no evidence that participants could distinguish between human and machine generated outputs. In other words, the approach based on abstracted Gaussian prototypes performed as well as possible: the generated characters were indistinguishable from human-made characters. Preference scores showed broad parity but with a significant difference from chance favoring the machine-produced outputs. In other words, judges tended to think the machine-generated examples were superior generative productions than those of humans. For additional detail, Figure \ref{fig:individual} provides a breakdown of the evaluations of each individual judge.

We compare against BPL with three criterion: (1) full BPL, (2) BPL lesioned with no learning-to-learn and (3) BPL with no compositionality. In (2), BPL normally has learning-to-learn: it uses experience from many past characters to learn good hyperparameters (priors/variability settings) so it can adapt quickly from one example. In the lesion, they disable that learned hyperparameter adaptation, so the model is effectively doing only token-level inference from the single training example. In (3) BPL normally represents a character as a composition of parts that can be reused and recombined. In this lesion, they compare BPL to a matched model that allows only one spline-based stroke for the whole character so the model can’t explain characters by assembling multiple stroke fragments; it’s forced into a single-piece template. We compare our generative results against the gold standard identification accuracy of BPL.

\begin{table}[h]
    \small
    \centering
    \caption{Descriptive Statistics for Average Scores Across Judges}
    \begin{tabular}{ccc}
         &  Identification & Preference\\
         Mean&  52.33\%& 55.33\%\\
         SD&  8.17\%& 8.27\%\\
 Min& 40.00\%&43.33\%\\
 Max& 63.33\%&70.00\%\\
    \end{tabular}    
    \label{tab:descriptives}
\end{table}

\textbf{Generating New Concepts from Type}
For the evaluation of generating new characters belonging to an alphabet, the identification accuracy across judges was (M = 52.00\%, SD = 14.73\%. Min = 40.00\%, Max = 90\%), p = 0.55. For comparison, BPL achieves 49\% and 68\%, for full BPL and no learning to learn lesion, respectively. The preference for the machine-made was (M = 49.00\%, SD = 15.18\%, Min = 20.00\%, Max = 80.00\%), p = 0.77. 

\textbf{Generating New Exemplars}
For the set of images corresponding to generating new exemplars of a particular class, the average identification accuracy across judges was (M = 57.50\%, SD = 14.82\%. Min = 30.00\%, Max = 80\%), p = 0.04. BPL achieves an accuracy of 52\%, 80\%, and 65\% for full, no learning to learn, and no compositionality lesion, respectively. The preference for machine-made in this specific task was (M = 59.50\%, SD = 12.76\%, Min = 30.00\%, Max = 80.00\%), p = 0.01. 

\textbf{Generating New Concepts (Unconstrained)}
For the final task of generating entirely new concepts independent of alphabet, the identification accuracy across judges was (M = 47.50\%, SD = 12.01\%. Min = 30.00\%, Max = 70\%), p = 0.36. BPL achieves 51\%, 64\%, and 68\% for full, no learning to learn, and no compositionality lesion, respectively. The preference for the machine-made in this task was (M = 57.50\%, SD = 12.93\%, Min = 30.00\%, Max = 80.00\%), p = 0.02. 

\textbf{Overall Findings for the Generative Tasks}
The primary measure of identification accuracy is close to chance performance meaning that judges were largely unable to tell which characters were made by human or machine. Accordingly,  one can conclude that the model-generated characters were quite convincing. In terms of the preference measures, machine-made characters were more frequently preferred overall and in two of the three generative subtasks. This could merit further exploration in terms of AI-generated content.

\section{Discussion}
It should not be lost from view that the present results fall short in some respects of the human-like performance reported by Lake et al. (2015) with a Bayesian program language (BPL) approach. The critical thing to make clear is the important advance made here. Lake et al. (2019) are emphatic about the shortcomings inherent in proposed approaches to the Omniglot challenge. First, they note that models which succeed on one-shot classification learning are not capable of performing the generative tasks and in many cases rely on ways of amending the classification task to make it easier. Our approach is the first that we are aware of which shows good (albeit not excellent) classification performance without altering the one-shot task specifications and in combination with strong generative task performance. Second, they note that the best approaches to generating new examples do not also perform the one-shot classification task and tend to produce "unarticulated strokes" or samples with either too little or too much variation; further these examples have not been tested using a visual Turing test which they are "doubtful they would pass." (Lake et al, 2019, p. 102). Once again, our approach is the first, to our knowledge, to show empirically-demonstrated BPL-level ability on the exemplar generative task. With regard to the task of constrained generation of new plausible concepts, Lake et al. (2019) found no competition for BPL, so our approach may be the first alternative to emerge. 

In theoretical terms, Lake and colleagues argue that the only viable path to success on the Omniglot challenge—with its demands for learning and generalization that are both systematic and flexible—comes via design principles of causality and compositionality. A further tenet they propose is that learning from scratch is not viable or human-like, i.e., that domain-general inductive biases and prior knowledge are a necessary foundation for learning the specific tasks in the Omniglot challenge. The argument for this point rests primarily on the empirical field of play: insofar as only BPL meets the challenge, then the point would appear to hold. However, the AGP approach presented here includes no causal component and passes only provisionally into what we call pseudo-structural elements that capture a loose generative model of character form (components and locations). Perhaps most dramatically, the strong performance demonstrated across discriminative and generative tasks is accomplished while holding the Omniglot challenge to a higher standard than its originators in that the AGP model relies on no additional information, learning, or learning how to learn: true one-shot means literally one shot and nothing else. Accordingly, the present results offer the first fully viable competition to BPL in terms of qualitative capabilities and directly challenges the implied message of Omniglot: we show how it is possible to succeed at a high level from scratch without relying on an already-in-place, general-purpose structured symbol system. 

A natural question is how the model captures relations among parts, for instance, that the two strokes of a "7" must meet at their endpoints, or that shifting one stroke should entail a corresponding shift in the other. Our model handles this implicitly rather than explicitly: because all k GMM components are estimated jointly from the same foreground coordinates in a shared spatial frame, the inferred means encode where each subpart sits relative to every other subpart. There is no relational grammar, but the spatial coherence of the character is preserved through joint estimation. This pseudo-structural sensitivity is also reflected in classification. The Tversky-based similarity metric penalizes spatial misalignment across the full AGP, meaning the model does not merely detect whether the right parts are present, but it is sensitive to whether those parts are in the right locations relative to each other. Inspection of classification trials confirms this: the model reliably rejects distractor characters that share similar stroke components with the target but arrange them in different spatial configurations. While purpose-built adversarial examples could likely expose limitations of this approach, the present results suggest that for naturalistic handwritten characters, the joint statistics of part locations carry sufficient relational information without requiring explicit symbolic encoding.

\section{Conclusion}
We present a novel approach for addressing the Omniglot challenge using Abstracted Gaussian Prototypes in conjunction with a VAE for generative tasks. AGPs leverage GMMs to build representative prototypes for each concept by abstracting the subparts of the single available instance for each class. The AGP functions as a generative image segmentation method applied to cluster the pixels of an image which offering a simple yet powerful method for encoding higher level representations that capture a quasi-structural model of 'what and where' for visual concepts from minimal data. One-shot classification is achieved by applying Tversky's set-theoretic similarity metric to compare AGPs. A novel AGP-VAE pipeline employs VAEs to utilize AGPs with different component numbers to generate diverse and compelling (i.e., consistent, yet original) variants that judges found indistinguishable from human drawings. 

While our approach represents a promising step toward true one-shot learning, there are several limitations to be acknowledged. First, we acknowledge that these are not state-of-the art results for one-shot classification; however, we are not aware of another approach that performs this well under the strict interpretation of one-shot learning as being fully from scratch. The classification performance should also be considered in the context of the model's excellent performance on the generative tasks in the challenge.  In terms of potential wider impact of our approach,  our method currently only handles line-based images with binary (on/off) pixels. The scalability to more complex natural images (i.e., ImageNet, CIFAR, etc.) would require new approaches to handle more features such as color. Furthermore, computational challenges from AGP generations and Tversky calculations can arise when working with non-binary images. Future research directions should aim to address these limitations and further refine the AGP framework for broader and more robust applicability in the field of one-shot learning. 

In sum, we present AGPs as a novel approach that achieves high performance on true one-shot classification task and proves impressively adaptability to the generative tasks of the Omniglot challenge. Critically, this degree and breadth of success is achieved without high model complexity, without slow and demanding computation, without lack of transparency, without the need for external pre-training, and without invoking a complex symbol system for explicit  structural recoding. Ideal takeaways from this work include: (1) the value of the computational cognition framework for productive crosstalk between cognitive science and ML, (2) the potential for approaches that are intermediate in nature between the poles of the statistical and symbolic frameworks, (3) the future potential of the design principles of the AGP and the AGP-VAE pipeline for true one-shot learning and generative tasks without having to first learn how to learn.

\printbibliography

\section*{Acknowledgements} We thank the members of the Learning and Representation in Cognition (LaRC) Laboratory at Binghamton University for their feedback during the pursuit of this project.

\section*{Appendix}

\textbf{Visual Turing Tests}\label{app}

The full set of questions from our "visual Turing tests" given to the human judges. The set of character outputs generated by our model are (from left to right, top to bottom): 

\begin{itemize}
  \item Within Class: A A B A B B A B B A 
  \item Same Class: B A A A B A B B A A 
  \item Unconstrained: A B B A A B A B A A 
\end{itemize}

\begin{figure}[h]
    \includegraphics[width=0.5\textwidth]{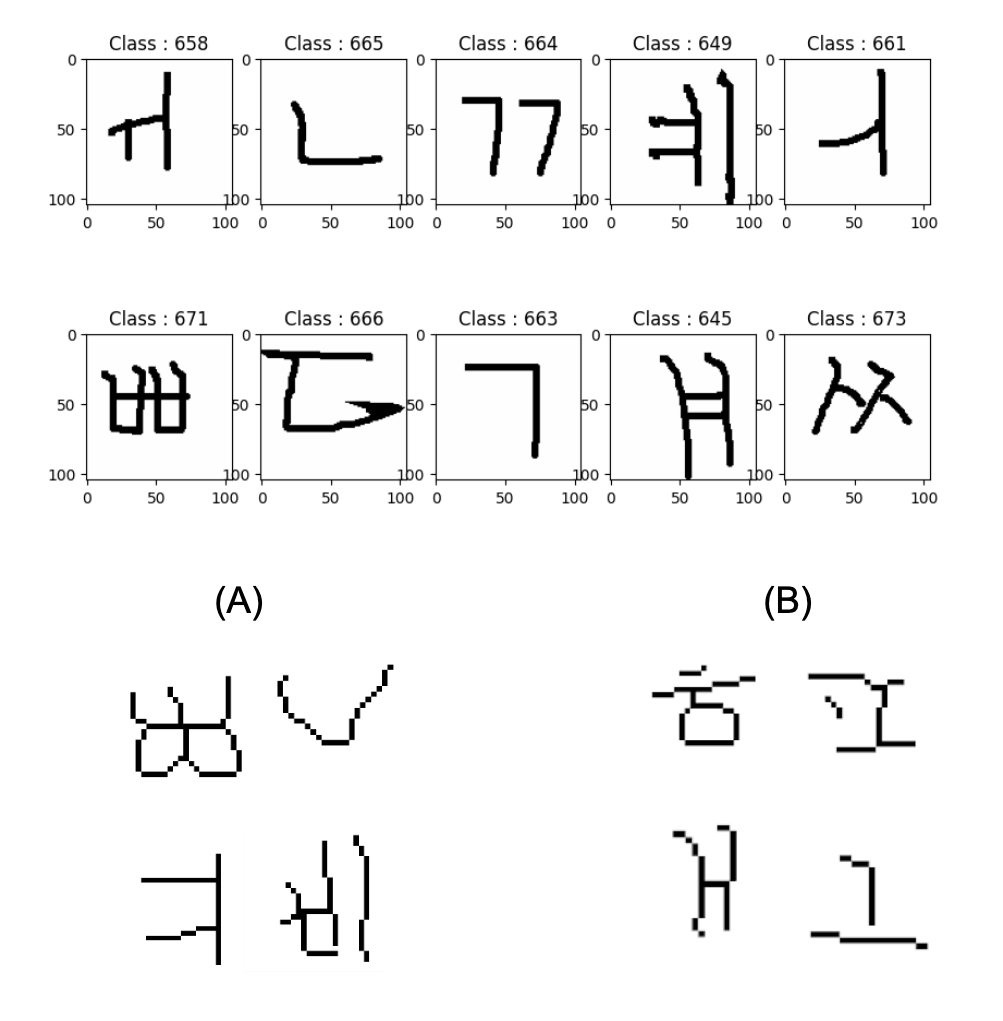}
    \includegraphics[width=0.5\textwidth]{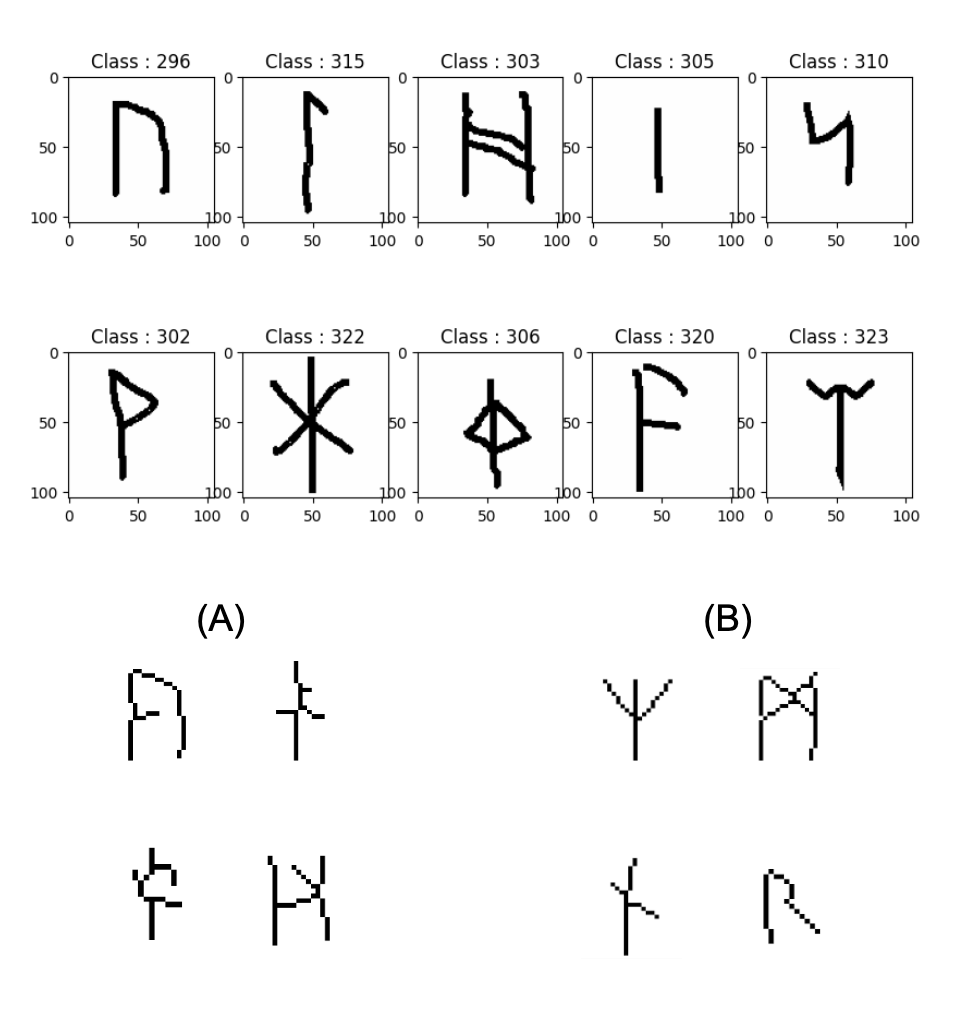} 
    \includegraphics[width=0.5\textwidth]{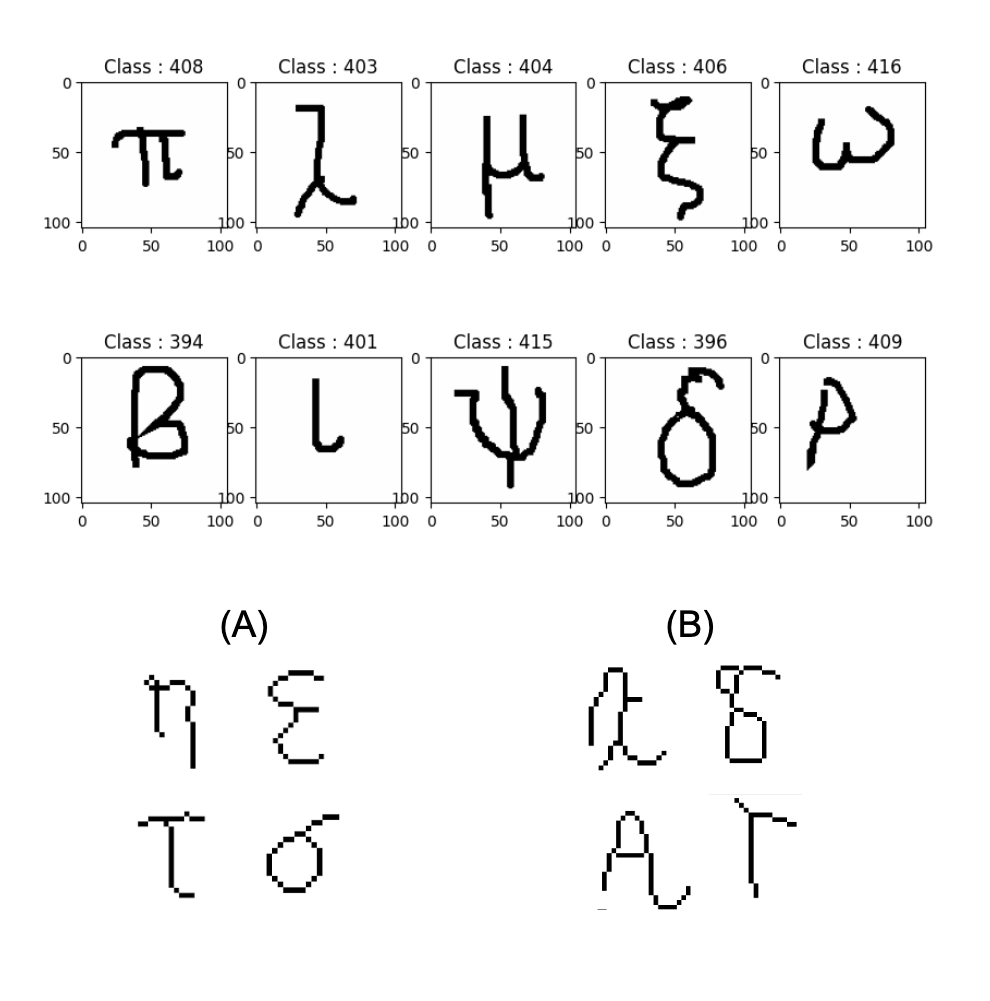}
    \includegraphics[width=0.5\textwidth]{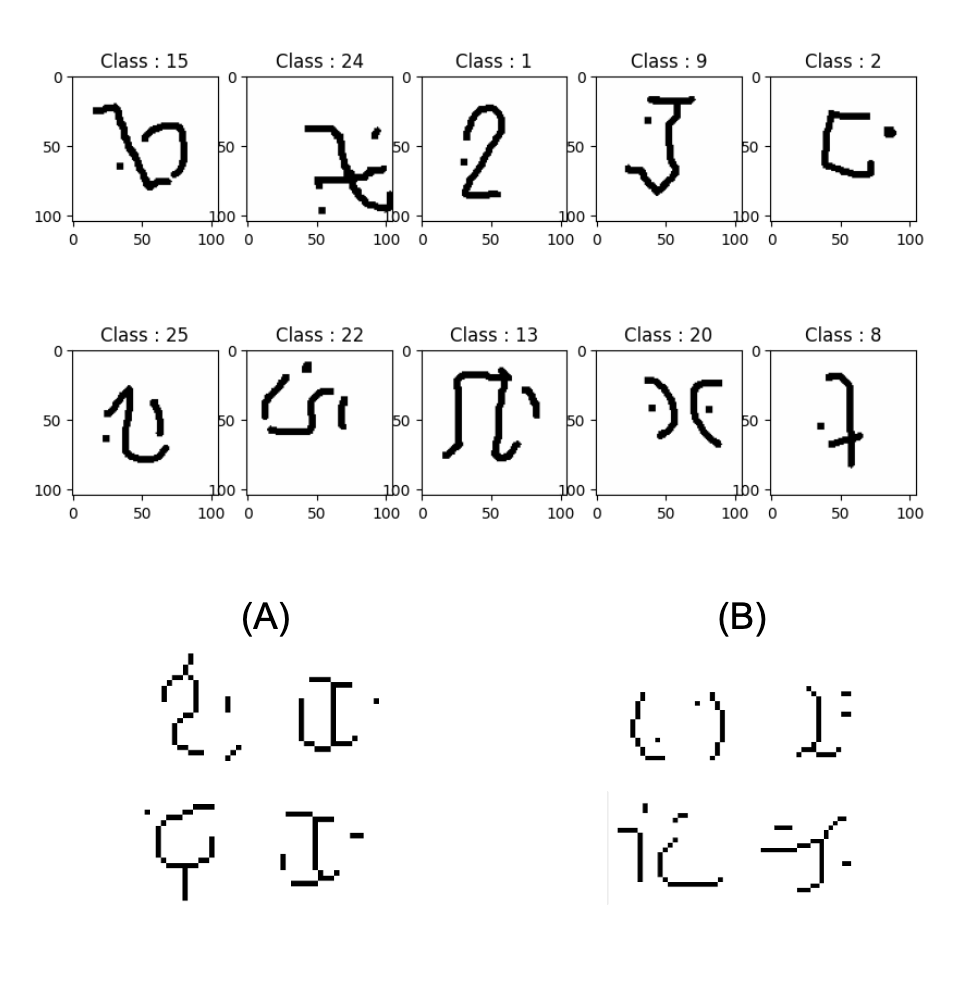} 
\end{figure}
\begin{figure}[h]
    \includegraphics[width=0.5\textwidth]{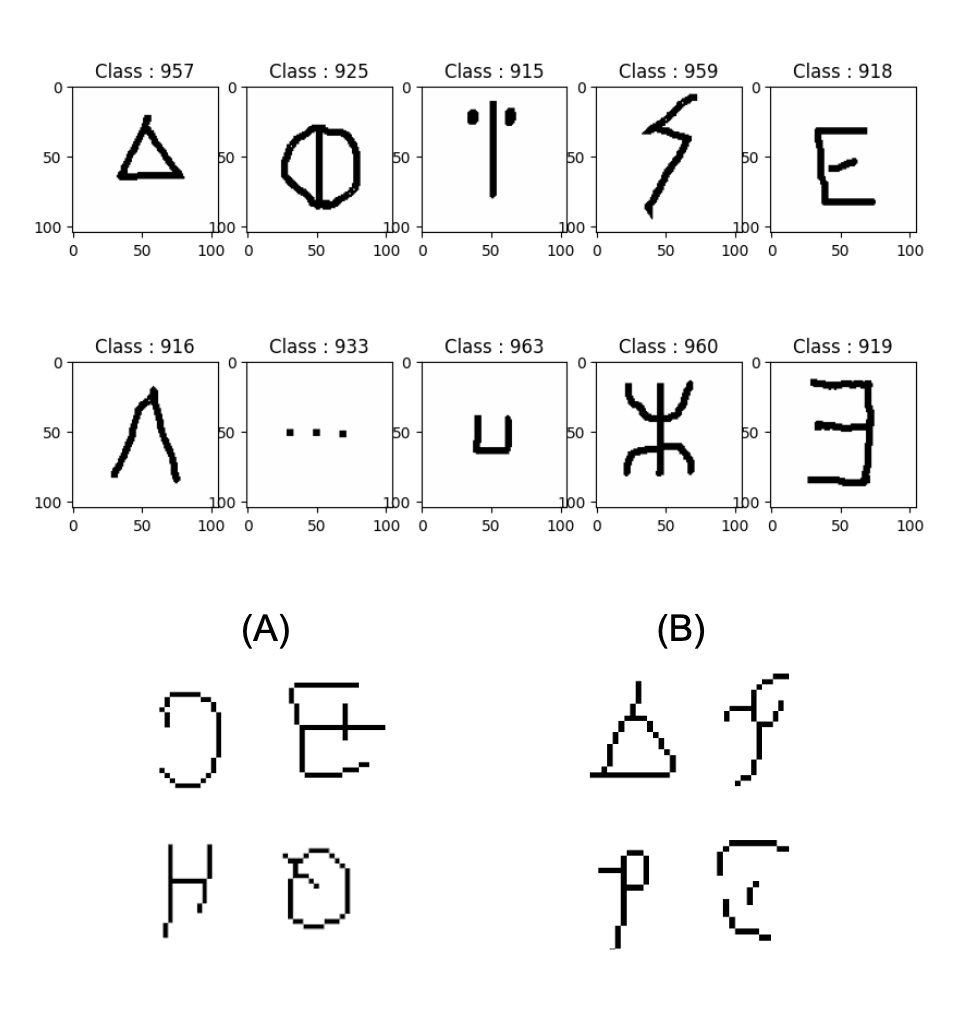}
    \includegraphics[width=0.5\textwidth]{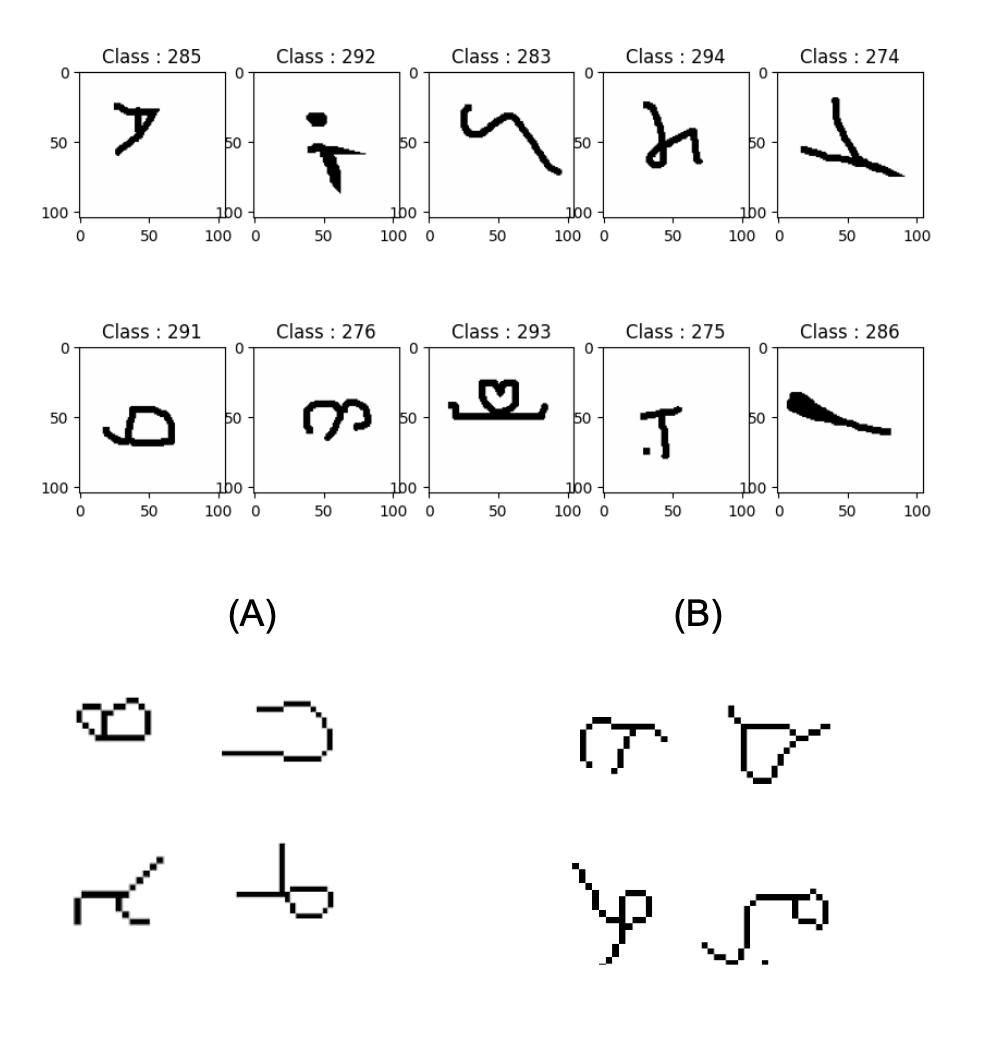} 
    \includegraphics[width=0.5\textwidth]{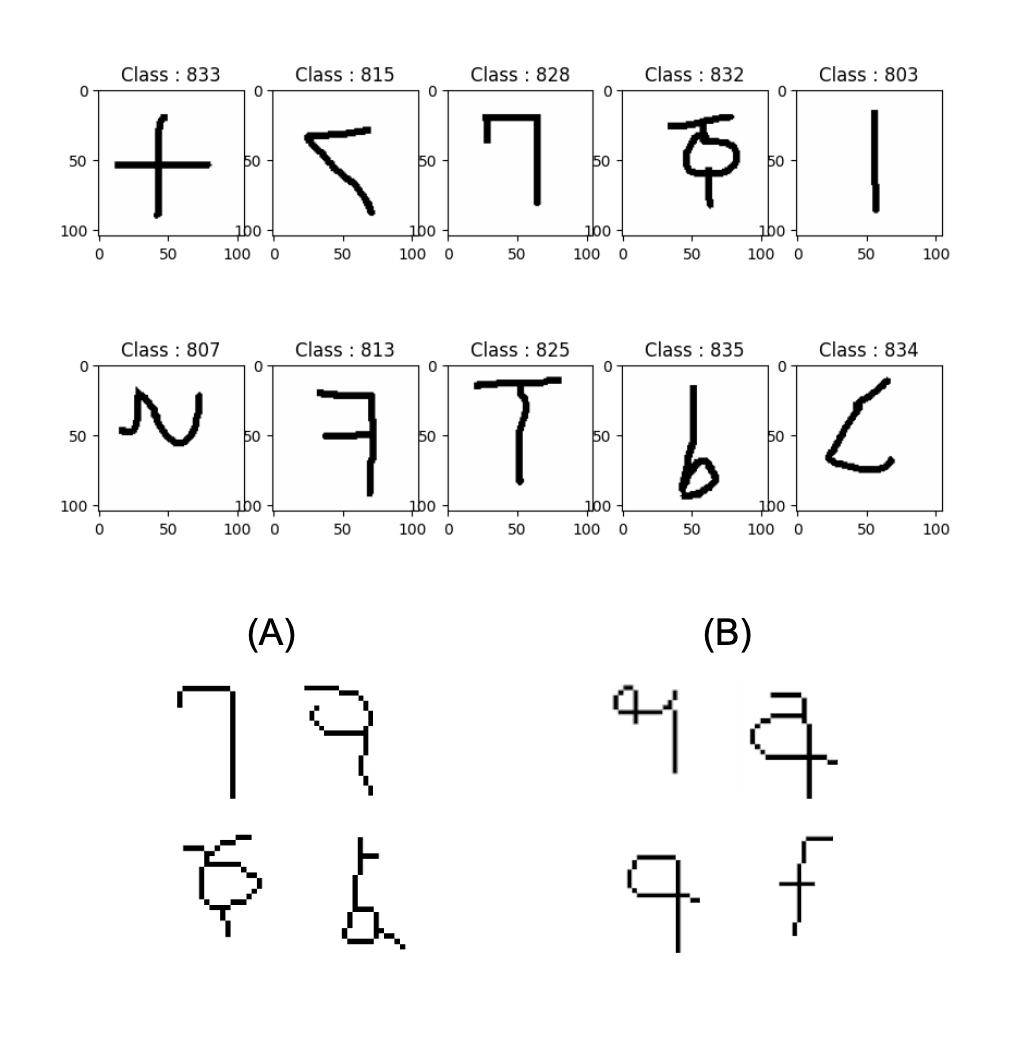}
    \includegraphics[width=0.5\textwidth]{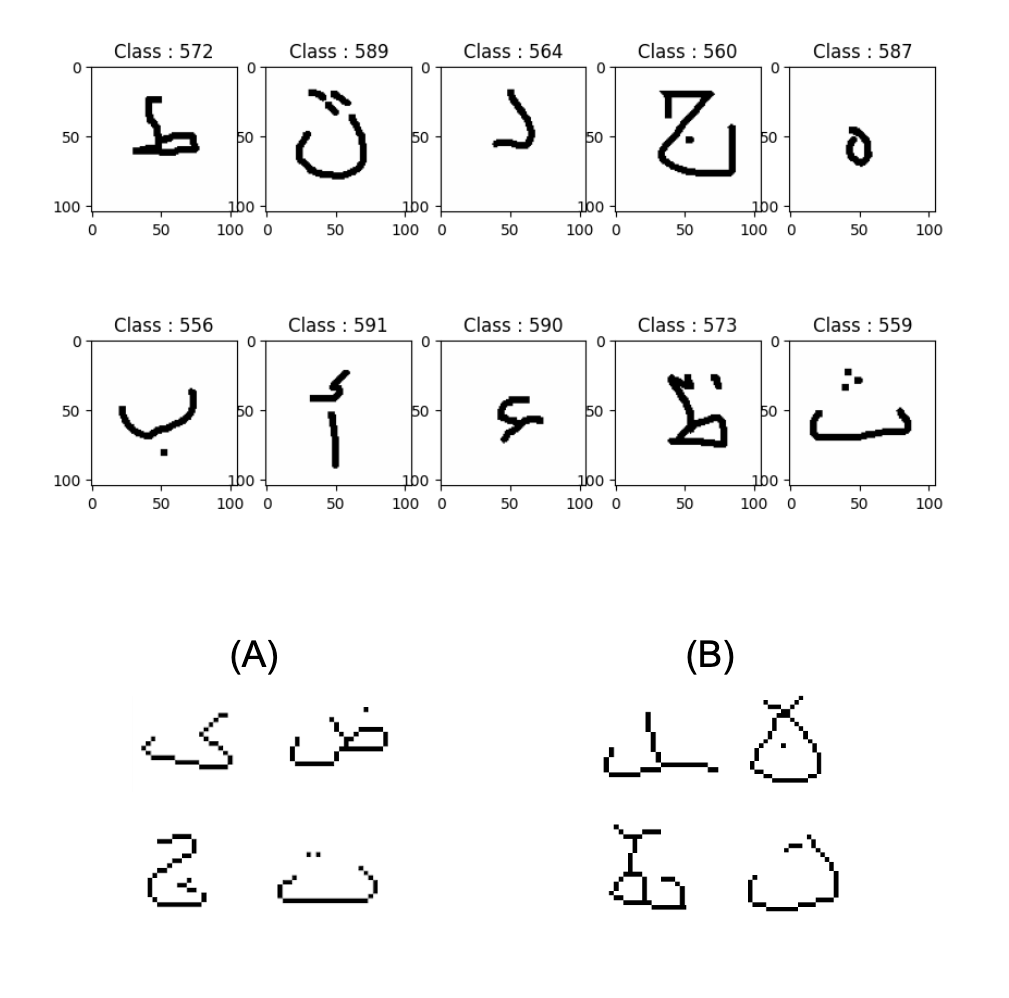} 
\end{figure}
\begin{figure}[h]
    \includegraphics[width=0.5\textwidth]{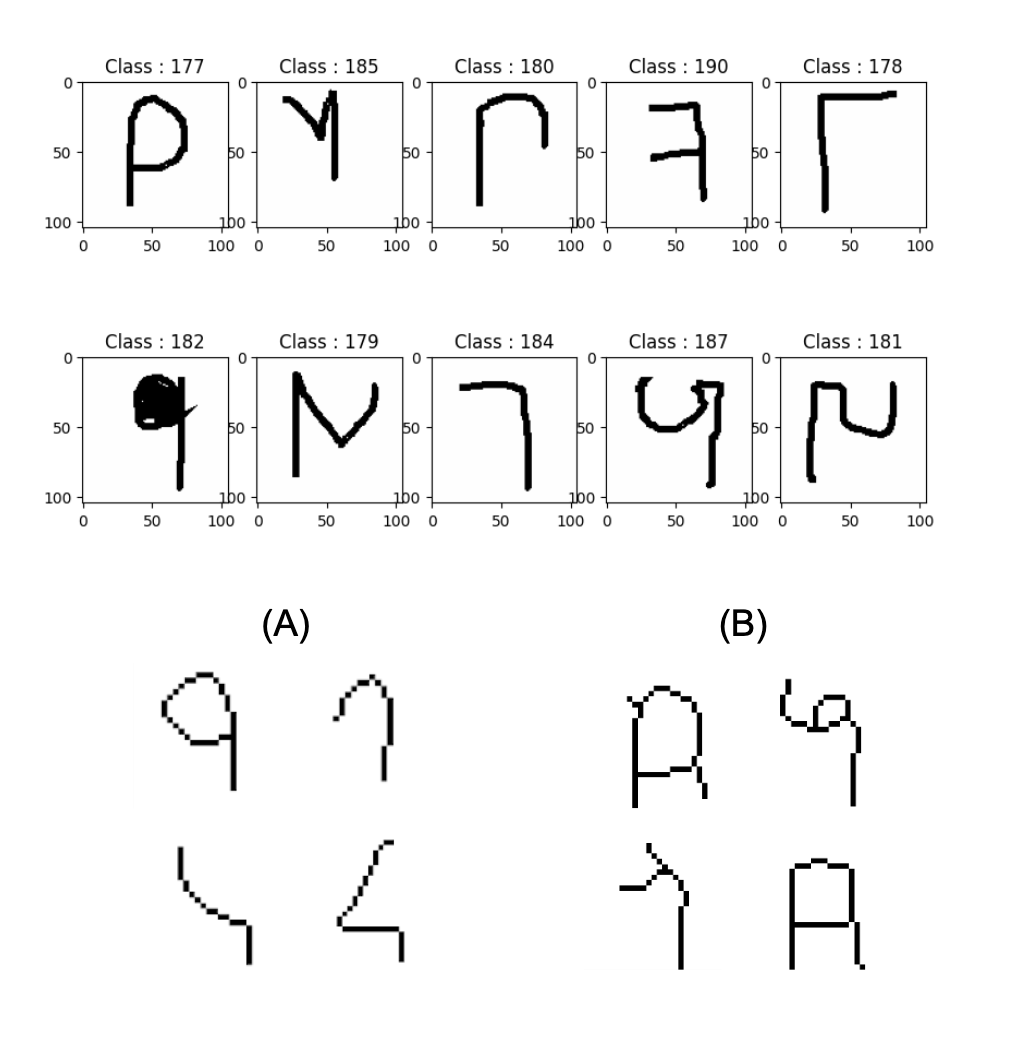}
    \includegraphics[width=0.5\textwidth]{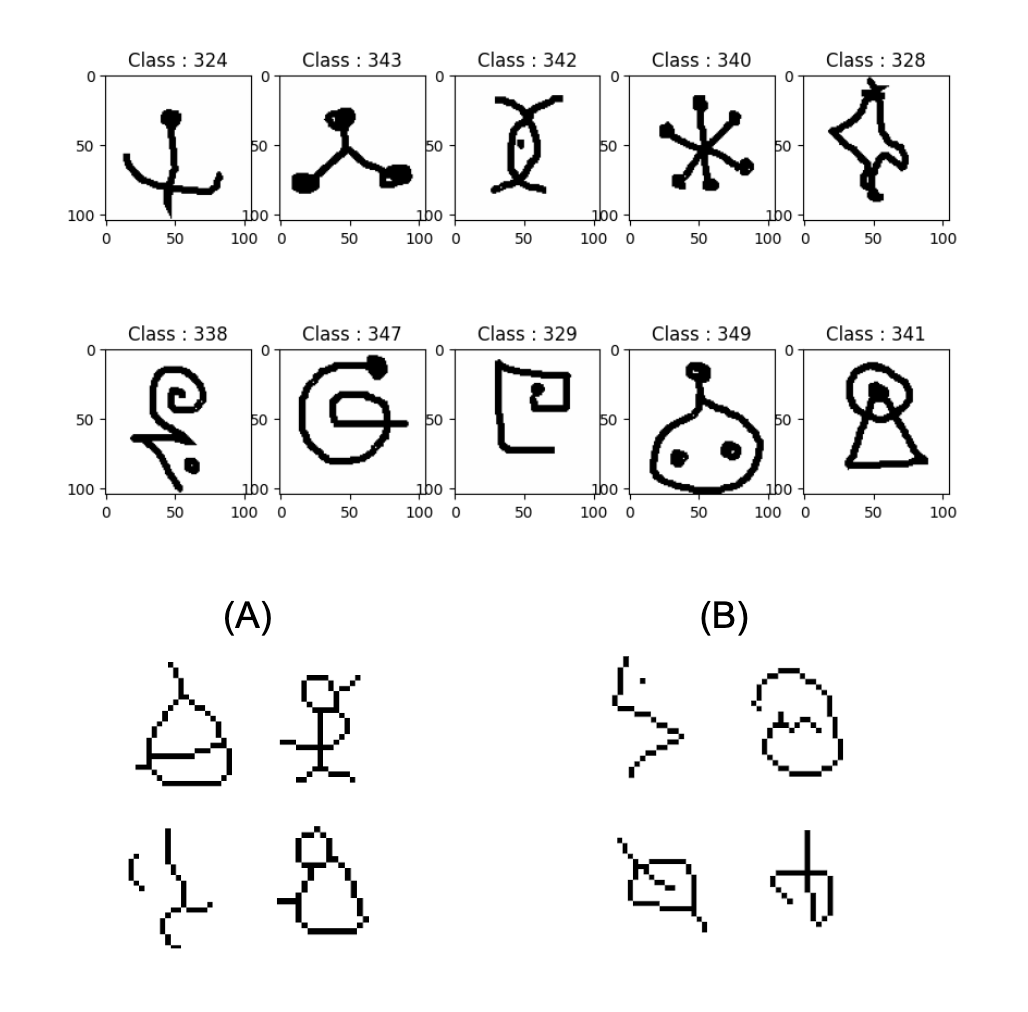} 
    \includegraphics[width=0.5\textwidth]{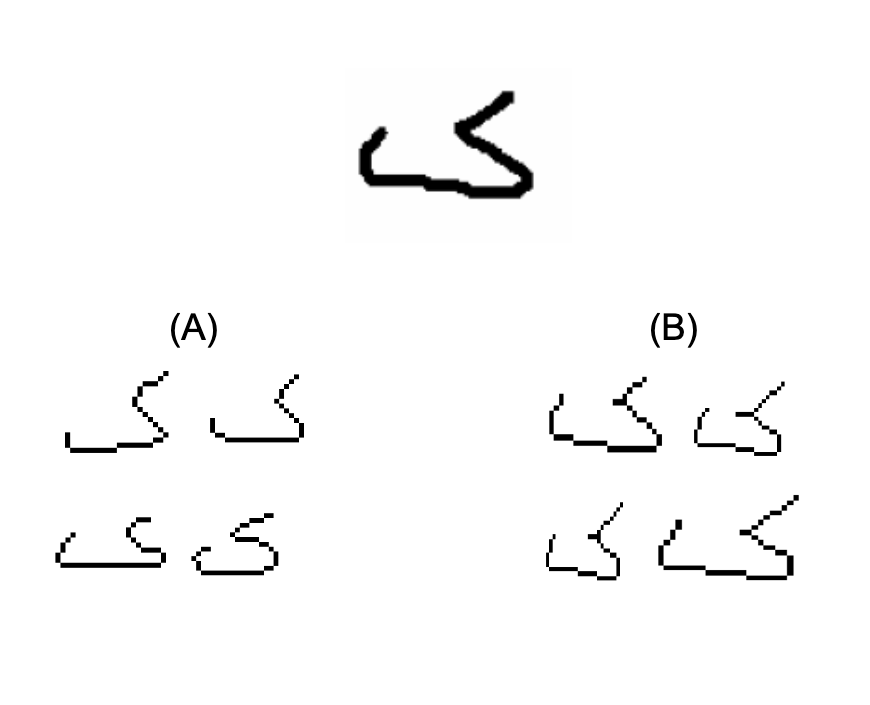}
    \includegraphics[width=0.5\textwidth]{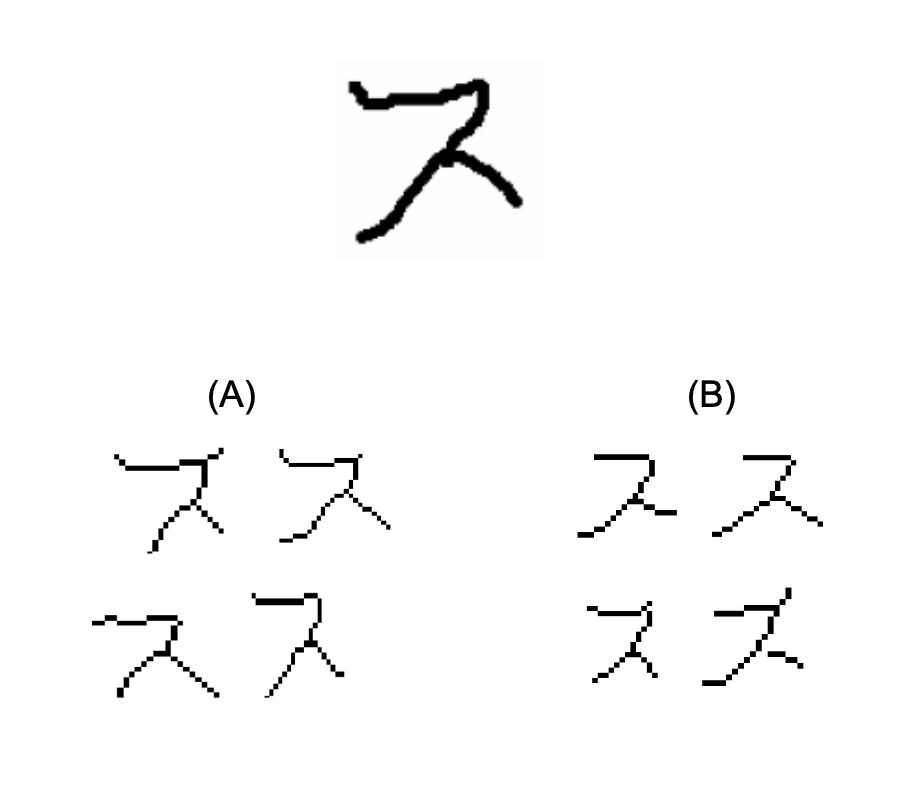} 
\end{figure}
\begin{figure}[h]
    \includegraphics[width=0.5\textwidth]{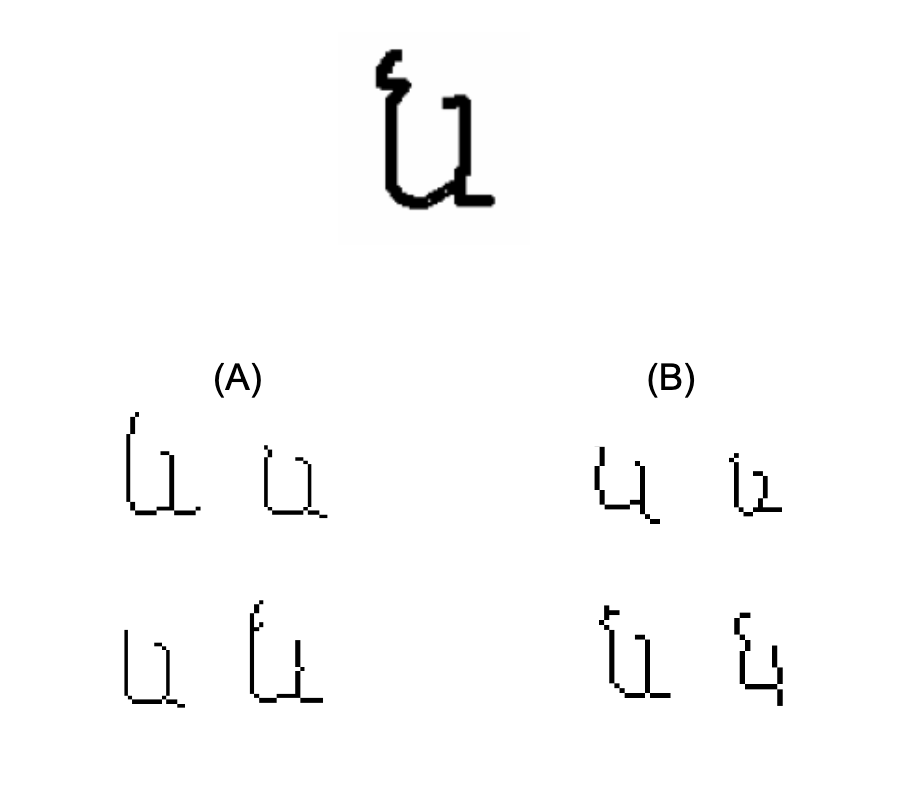}
    \includegraphics[width=0.5\textwidth]{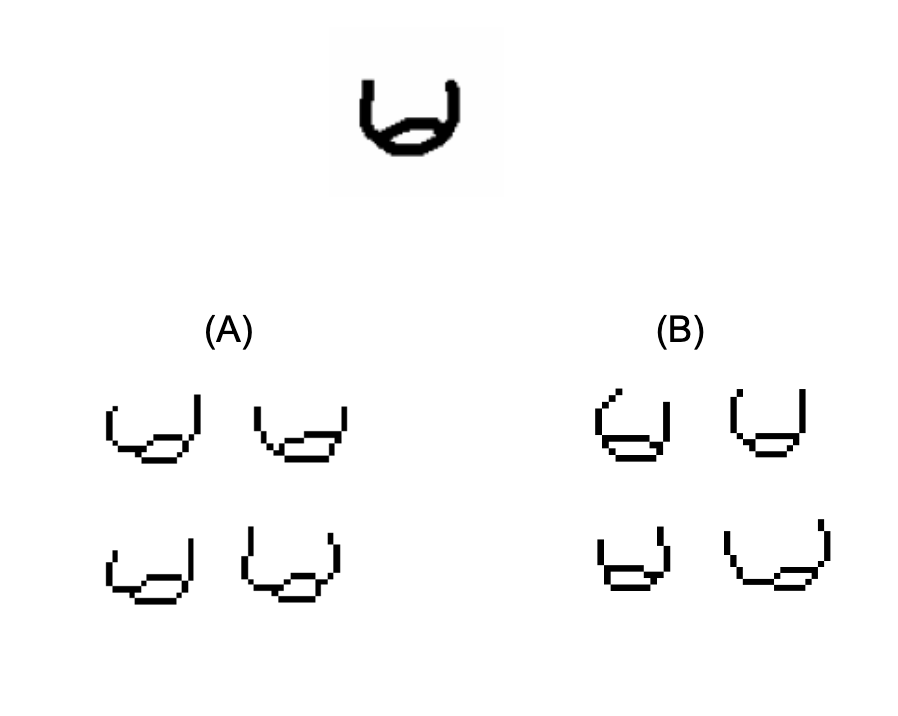} 
    \includegraphics[width=0.5\textwidth]{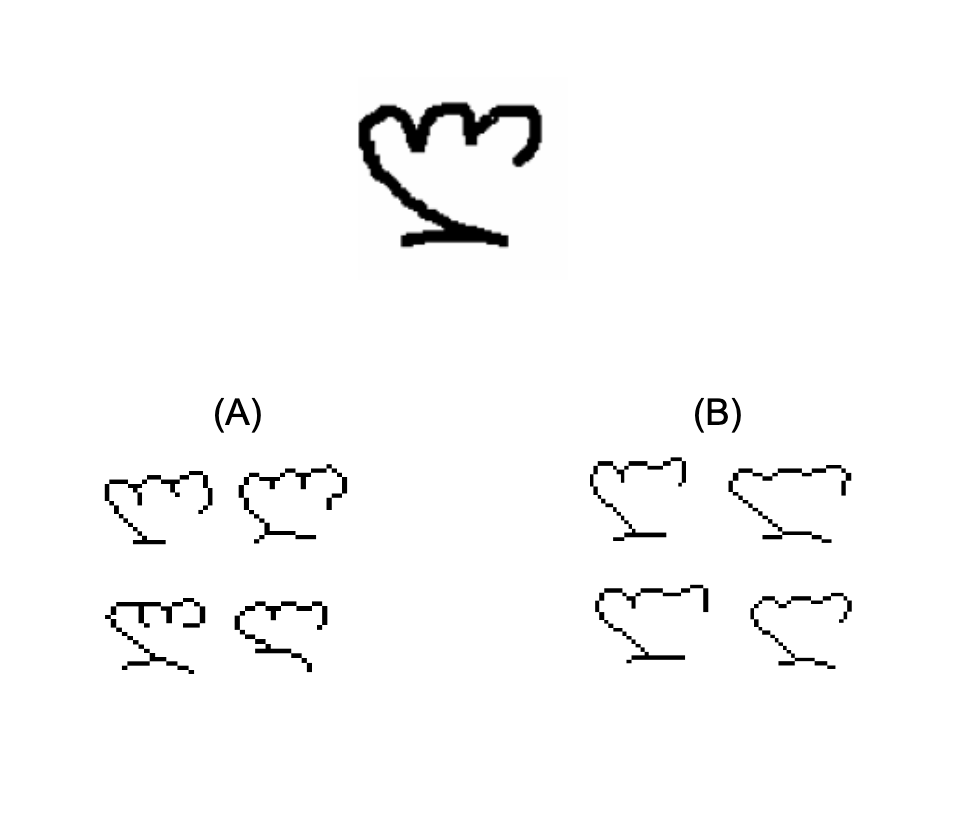}
    \includegraphics[width=0.5\textwidth]{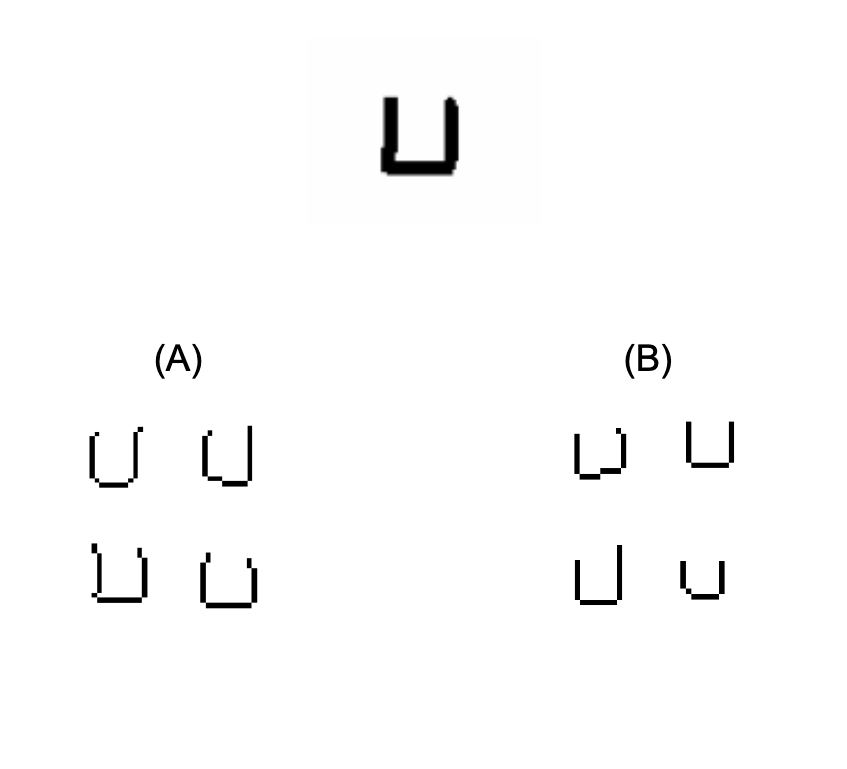} 
    \includegraphics[width=0.5\textwidth]{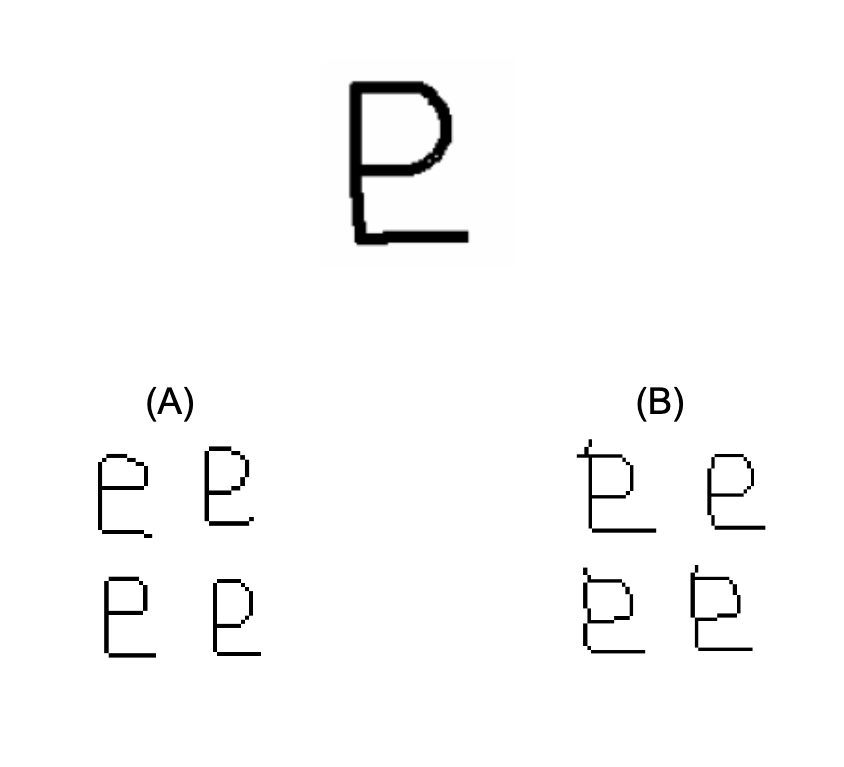}
    \includegraphics[width=0.5\textwidth]{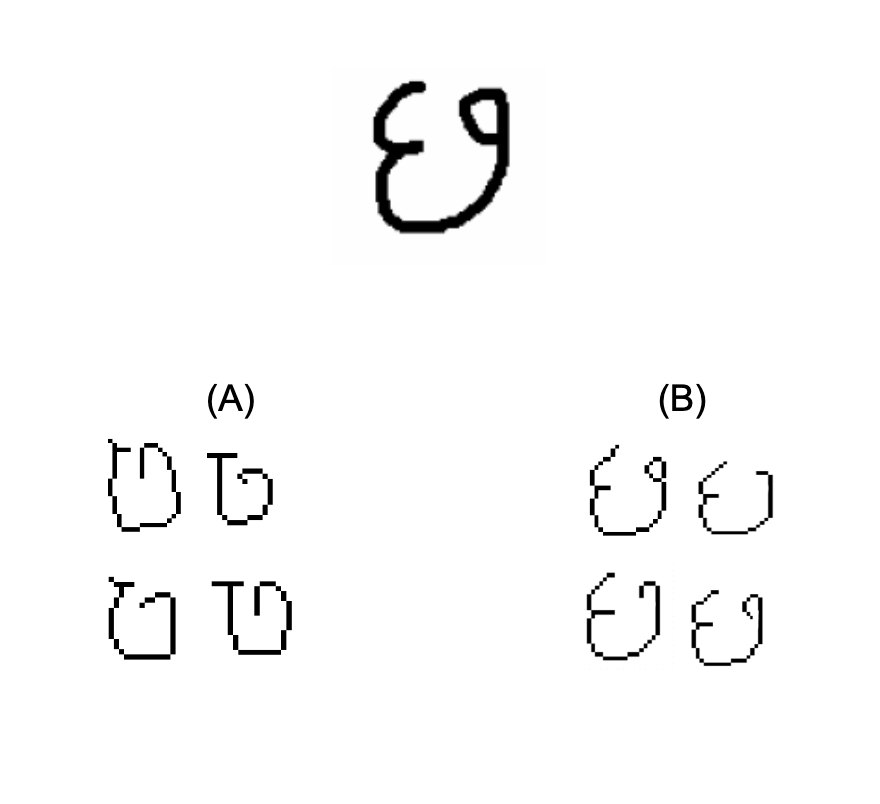} 
\end{figure}
\begin{figure}[h]
    \includegraphics[width=0.5\textwidth]{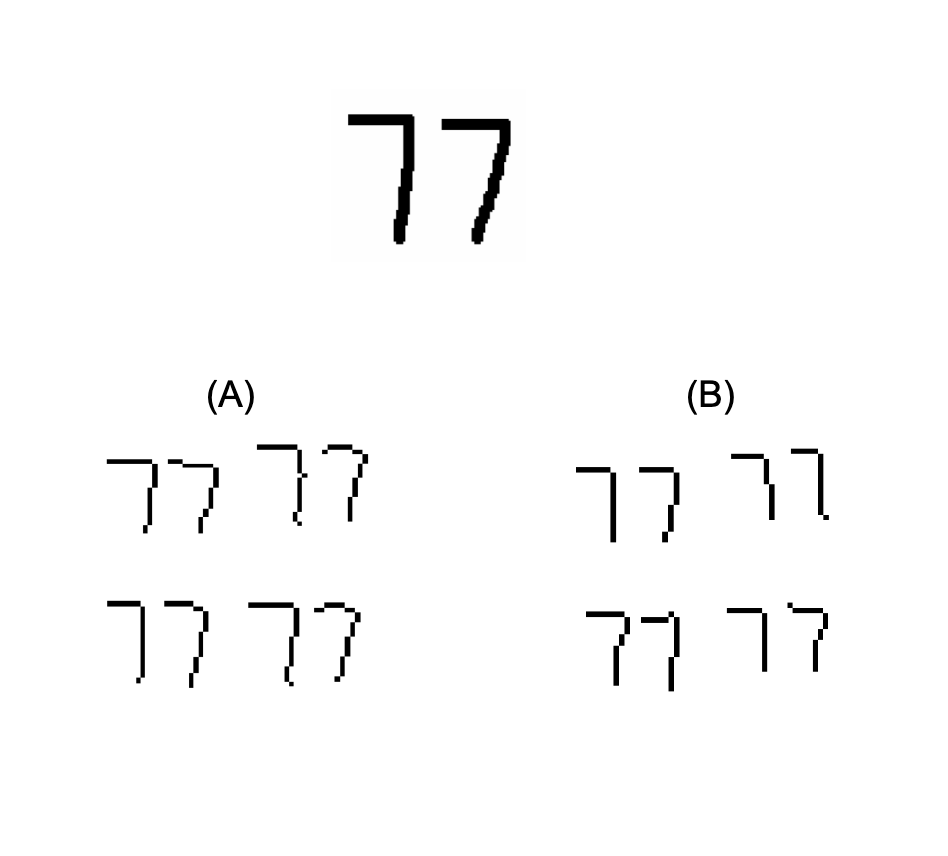}
    \includegraphics[width=0.5\textwidth]{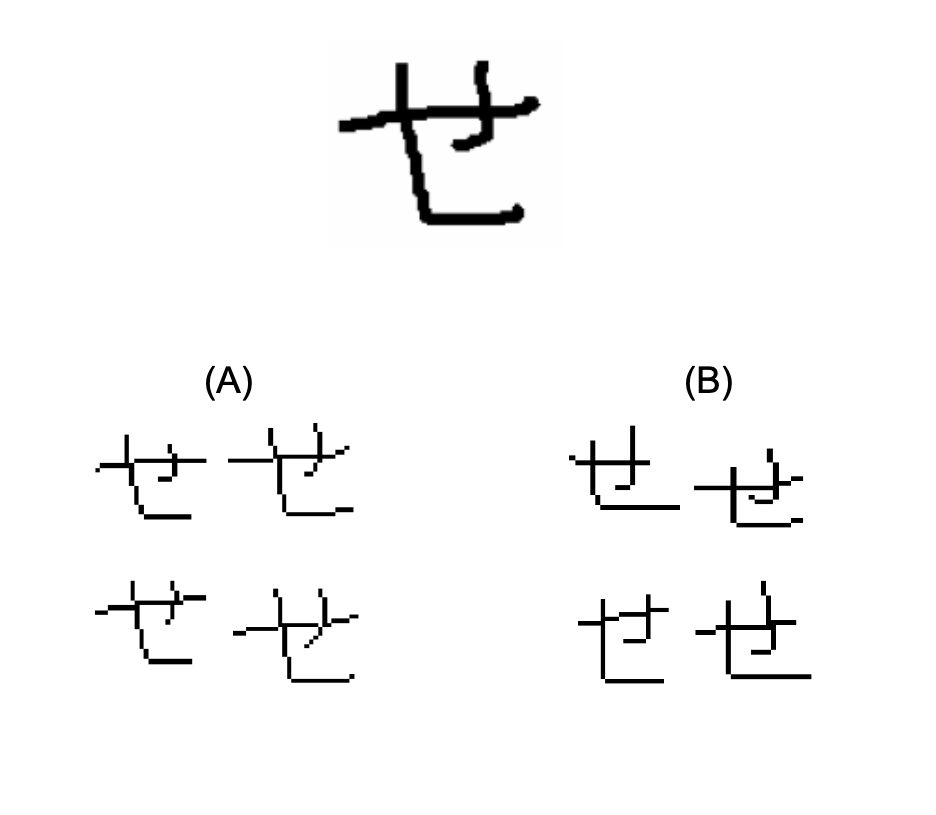} 
    \centering
    \includegraphics[width=0.6\textwidth]{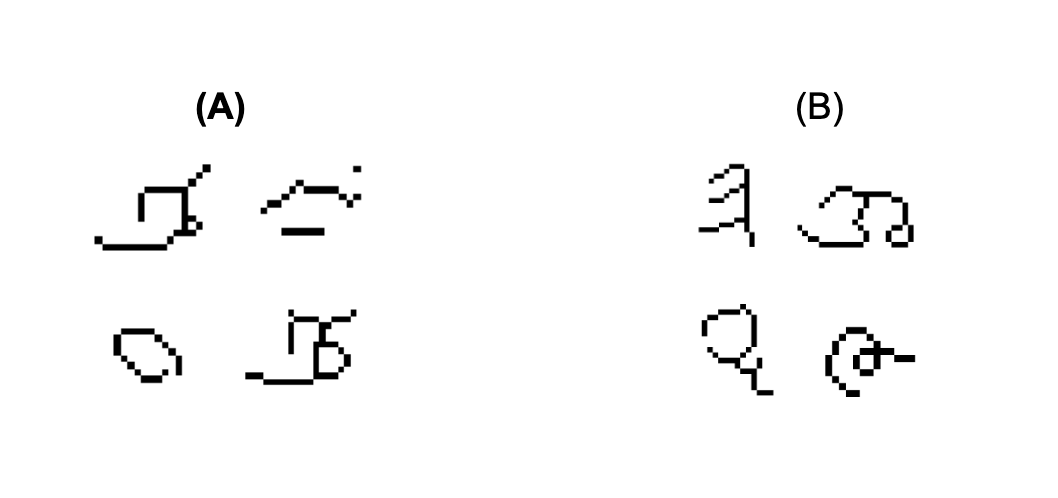}
    \includegraphics[width=0.6\textwidth]{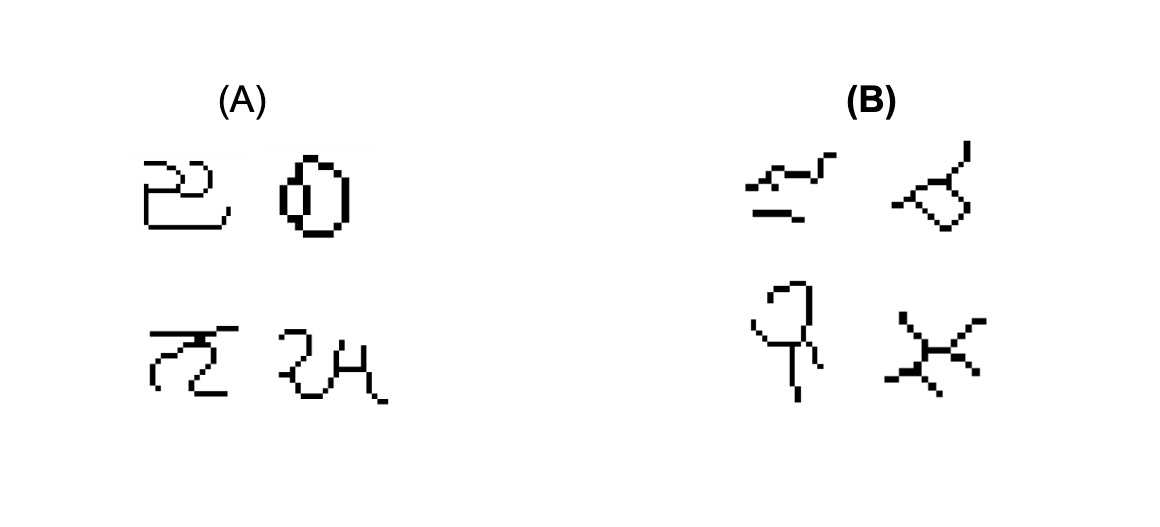} 
\end{figure}
\begin{figure}
    \centering
    \includegraphics[width=0.6\textwidth]{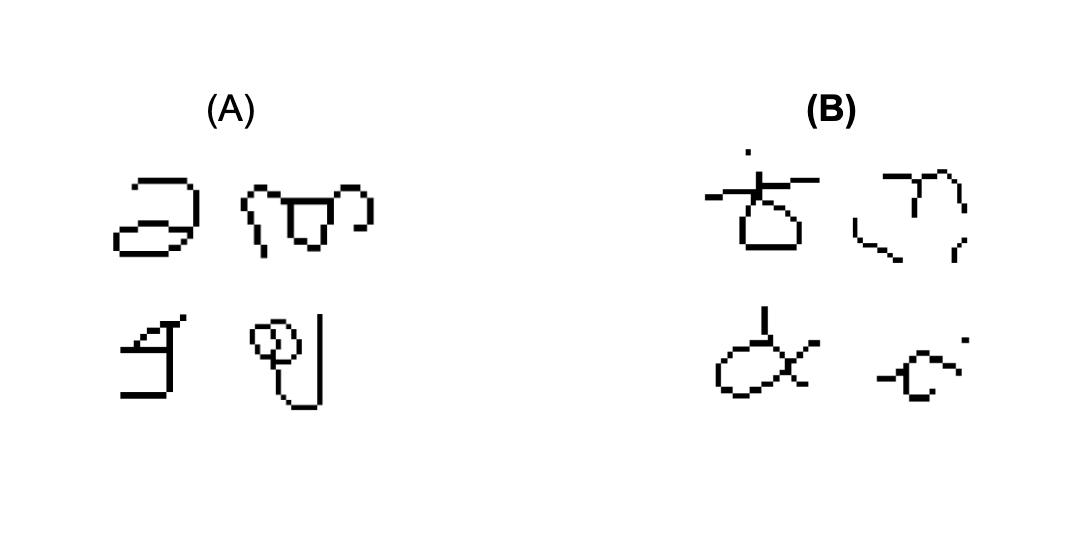}
    \includegraphics[width=0.6\textwidth]{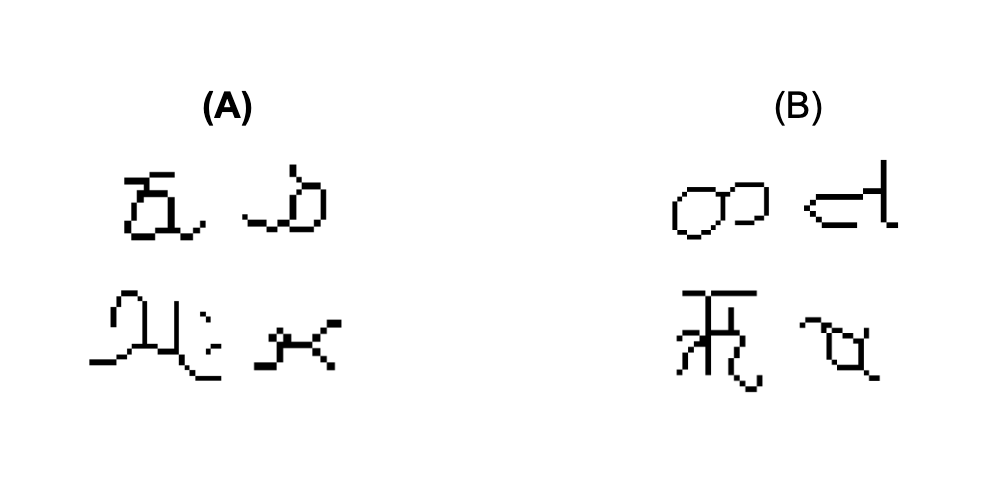} 
    \includegraphics[width=0.6\textwidth]{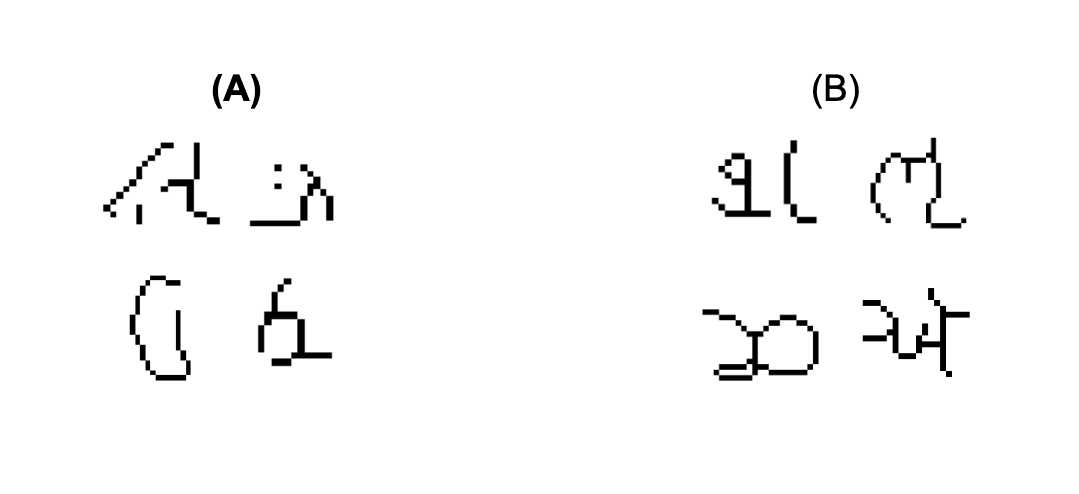}
    \includegraphics[width=0.6\textwidth]{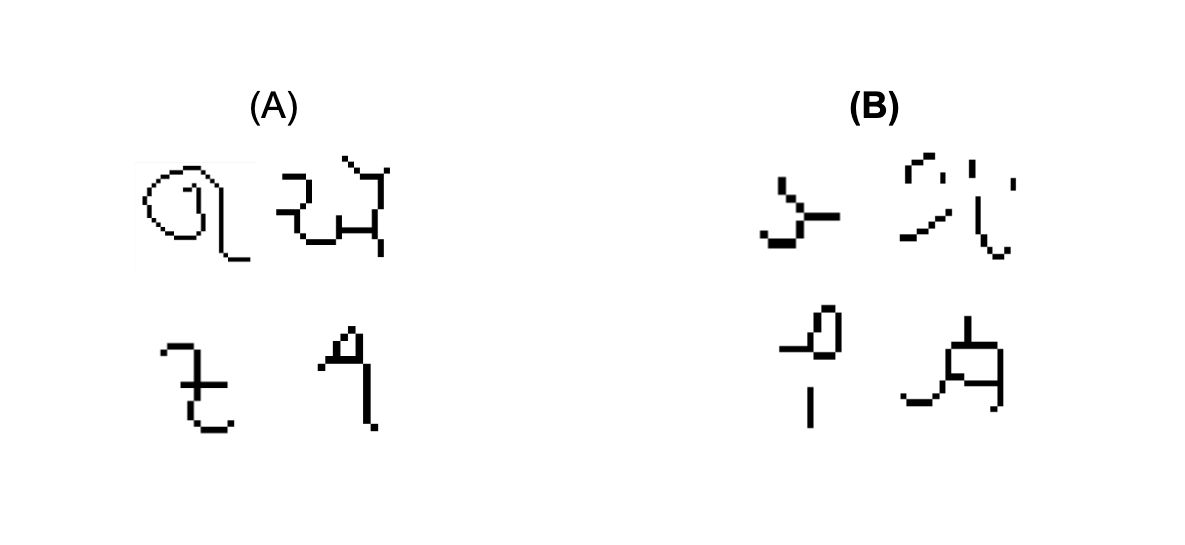}
\end{figure}
\begin{figure}
    \centering
    \includegraphics[width=0.6\textwidth]{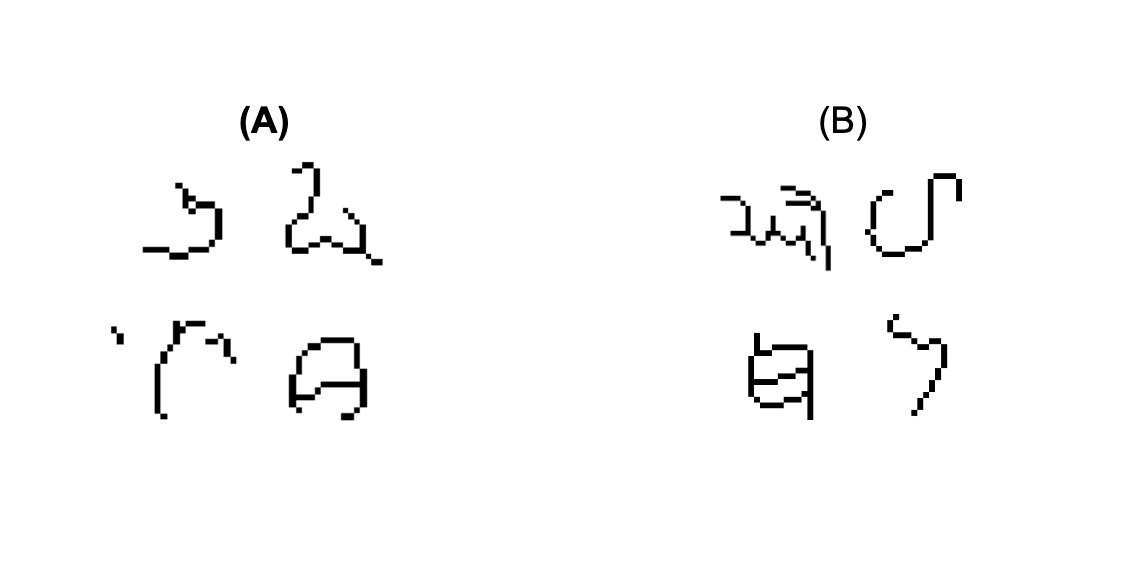} 
    \includegraphics[width=0.6\textwidth]{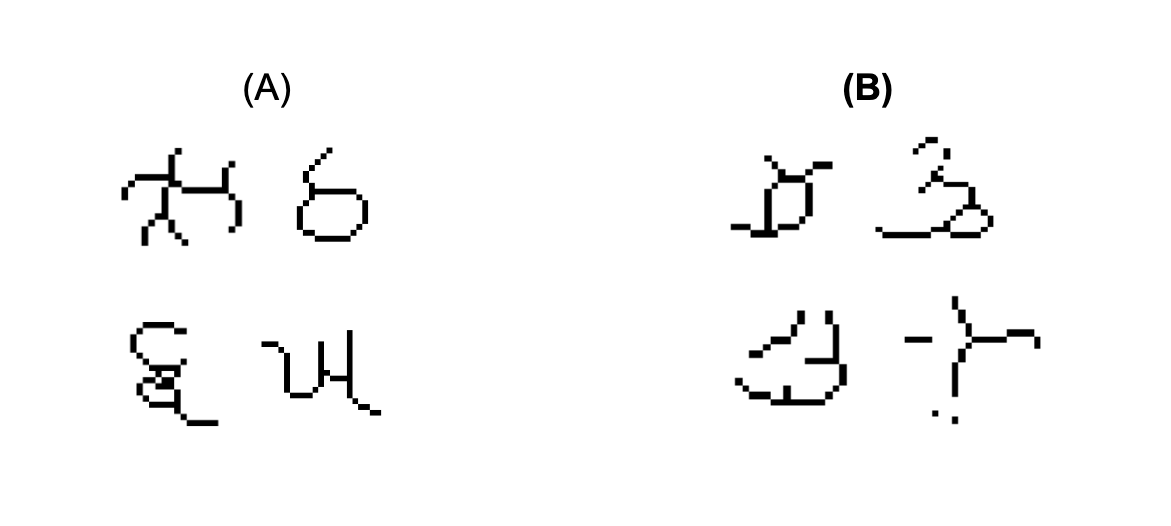}
    \includegraphics[width=0.6\textwidth]{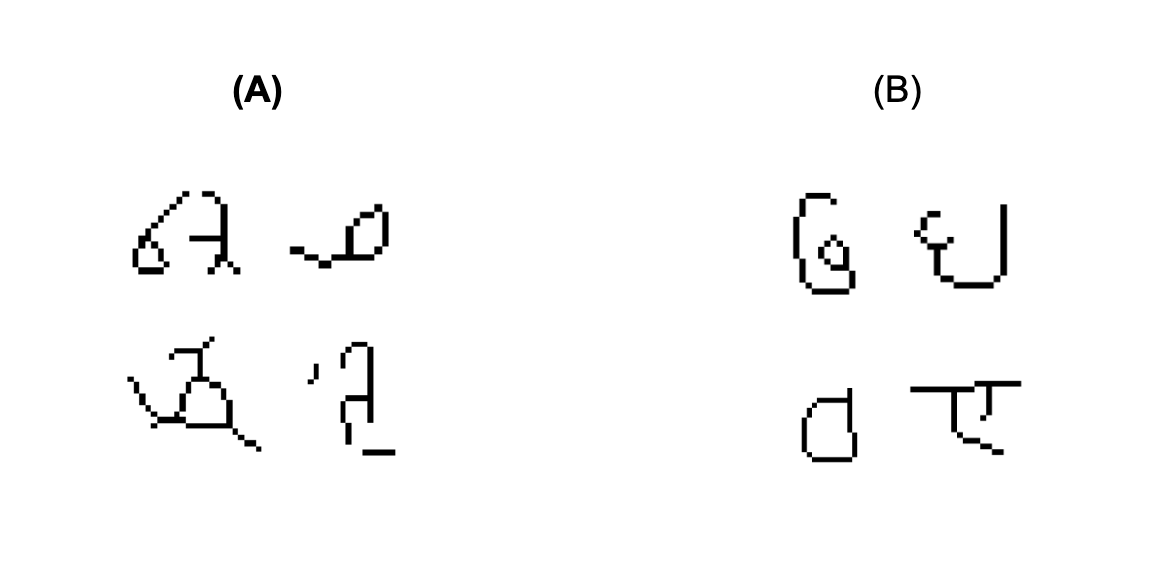}
    \includegraphics[width=0.6\textwidth]{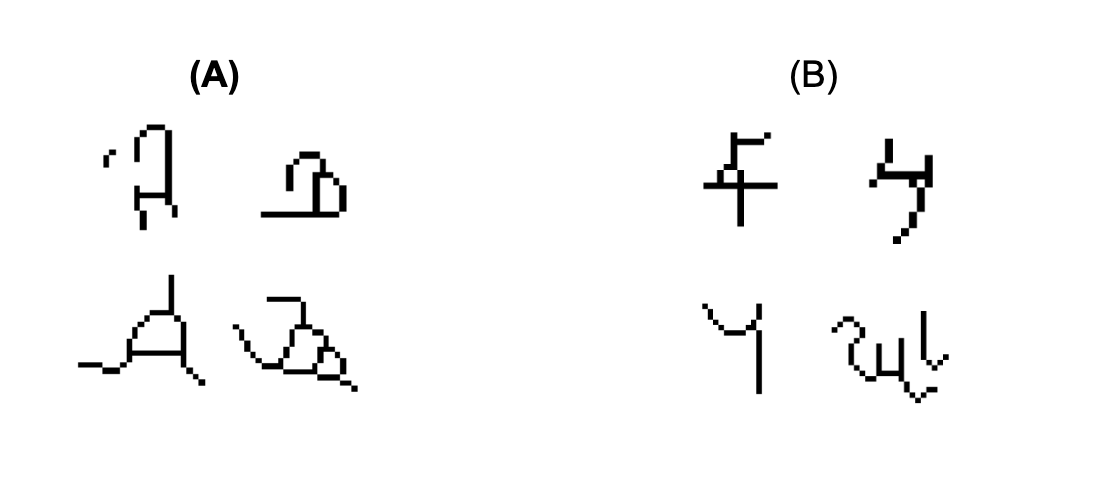} 
\end{figure}

\end{document}